\documentclass{iopjournal}

\usepackage{amsmath,amsfonts}
\usepackage[svgnames]{xcolor}
\usepackage{graphicx}
\usepackage{multirow}
\usepackage[numbers,sort&compress]{natbib}
\usepackage{amssymb}
\usepackage{tikz}
\usepackage{placeins}
\usepackage[switch]{lineno}
\usepackage{orcidlink}
\usepackage{fancyhdr}

\definecolor{darkbrown}{rgb}{0.4,0.26,0.13}
\definecolor{darksienna}{rgb}{0.24,0.08,0.08}
\definecolor{darkpowderblue}{rgb}{0.0,0.2,0.6}

\usetikzlibrary{shapes.geometric,arrows.meta,positioning,calc,fit,backgrounds}

\hypersetup{
    linkcolor=darkpowderblue,
    citecolor=cyan,
    urlcolor=darksienna
}

\let\orcid\orcidlink

\begin{document}


\title{Transformers with Physics-Informed Encodings and Simulation-Based Inference for Robust Detection of Eccentric Binary Black Holes in Pulsar Timing Array Data}

\author{
Subhajit Dandapat$^{1}$\orcid{0000-0003-4965-9220},
Alvin J.~K.~Chua$^{1,2}$\orcid{0000-0001-5242-8269}
}

\affil{$^1$Department of Physics, National University of Singapore,
21 Lower Kent Ridge Road, Singapore 119077}

\affil{$^2$Department of Mathematics, National University of Singapore,
Singapore 119076}

\email{subhajit.phy97@gmail.com}


\begin{abstract}
Pulsar timing arrays (PTAs) provide a unique window into nanohertz gravitational waves (GWs), but extracting astrophysical parameters from noisy, long-baseline timing residuals remains computationally challenging with traditional Bayesian techniques due to the high dimensionality of the parameter space, complex and correlated noise models, and the cost of repeated likelihood evaluations. We introduce a Transformer with a physics-informed positional-encoding framework for the efficient inference of eccentric binary black holes in relativistic orbits from PTA data. Our approach embeds analytical GW phase evolution directly into the model through structured positional encodings, enabling the network to learn physically meaningful representations from raw PTA timing residuals. We then use generative models, including discrete and continuous conditional normalizing flows, to infer posterior distributions within a simulation-based inference framework. Across a range of signal-to-noise ratios, the proposed method achieves improved accuracy, sharper posteriors, and faster inference compared to physics-agnostic baselines. While presented for deterministic white-noise signals, the modular framework readily generalizes to realistic PTA analyses incorporating red noise and additional components. This work highlights the potential of physics-aware deep learning models as scalable alternatives to conventional inference pipelines for next-generation PTA datasets.
\end{abstract}


\section{Introduction}

The Pulsar Timing Array (PTA) aims to detect nanohertz gravitational waves (GWs) and is considered a crucial complement to the detection of GWs with laser interferometers. This PTA-based approach represents a major milestone in GW astrophysics, following the detection of GWs in the Laser Interferometer Gravitational-Wave Observatory (LIGO) frequency band~\citep{LIGOScientific:2016aoc}.  Recent independent and coordinated efforts from North American Nanohertz Observatory for Gravitational Waves (NANOGrav:~\citep{agazie2023ng15yrgwb}), the European Pulsar Timing Array and the Indian
Pulsar Timing Array (EPTA+InPTA:~\citep{wm3_einpta_23}), the Parkes Pulsar Timing Array (PPTA:~\citep{reardon2023pptadr3gwb}), and the Chinese Pulsar Timing Array (CPTA:~\citep{xu2023cptagwb}) have provided strong evidence for the existence of a nanohertz (nHz) stochastic gravitational wave background (SGWB) in their respective datasets. The International Pulsar Timing Array (IPTA:~\citep{perera2019international,agazie2023ipta3p+}) consortium is currently combining a data set of $\sim$ 115 MSPs of various PTAs, aiming to detect SGWB with higher significance. These experiments exploit the remarkable timing stability of millisecond pulsars to search for quadrupolar-correlated deviations in pulse arrival times, indicative of GWs. A well-known source in this frequency range is a population of sub-parsec–separated inspiralling supermassive black hole binaries (SMBHBs), whose individual GW signals combine incoherently to form a stochastic background~\citep{phinney2001practical}.
As PTA data volumes continue to grow, PTAs will become sensitive to GWs from isolated inspiraling SMBHBs, which is expected to happen in the coming years.
While PTAs have yet to achieve a firm detection, they have placed increasingly stringent constraints on the presence of such sources~\citep{ind_agazie2023nanograv,ind_antoniadis2024second}.

Despite steady progress in PTA science, conventional data-analysis approaches, particularly MCMC-based inference, struggle to meet the demands of current datasets~\citep{babak_forecasting_2024,van_haasteren_measuring_2009,enterprise,justin_ellis_2017_1037579}. PTA observations span long time baselines and involve many coupled astrophysical and noise parameters, creating an inference problem that is both high dimensional and computationally heavy~\citep{taylor_searching_2013}. As a result, likelihood evaluations in either the time or frequency domain can become prohibitively expensive. In practice, MCMC methods often converge slowly, mix poorly, and require extensive computational resources to obtain reliable posteriors~\citep{falxa_modeling_2024,freedman_efficient_2023}. Parameter correlations and degeneracies further complicate the exploration of the space, sometimes leading to biased or weakly constrained estimates~\citep{van_haasteren_new_2014}. As PTA datasets continue to expand in scope and complexity, these challenges highlight the need for more scalable and robust techniques for parameter estimation~\citep{hobbs_international_2010}.
With the advancement of intelligent data-processing techniques, machine
learning has become an increasingly powerful tool in GW data analysis.
Several studies have demonstrated that ML-based approaches can significantly
reduce computational costs while maintaining sensitivity comparable to
traditional pipelines. These methods have been applied to signal
detection~\citep{George:2017vlv,Gabbard:2017lja,Nousi:2022dwh}, glitch
classification~\citep{Chatterjee:2024ddw}, parameter estimation~\citep{chua2020learning,Dax2021real,Gabbard2022bayesian,Dax2025RealTime}, and other inference problems across the broader GW domain~\citep{cuoco2025applications}.
In the PTA sector, ML-based studies have investigated point-parameter
estimation for eccentric SMBHBs using simulated PTA residuals
\citep{chen2020machine}, showing that feed-forward and recurrent
architectures can recover chirp-mass--related quantities under simplified
noise assumptions.

Beyond point estimates, there has been growing interest in
simulation-based inference (SBI)~\citep{liang2025recent} for PTA data analysis, aimed at enabling
near-instantaneous generation of full posterior distributions. In
particular, normalizing flows (NFs)
\citep{kobyzev2020normalizing,papamakarios2021normalizing} have been used
to accelerate inference by several orders of magnitude. Their application to
PTA datasets has shown that, once trained, flow models can generate posterior
samples in seconds while maintaining accuracy, as demonstrated in analyses
using the subset of pulsars that contributed most strongly to the evidence
for a SGWB~\citep{shih2024fast}. Additional
developments include stochastic-gradient variational Bayes with
NFs~\citep{vallisneri2025rapid}, which directly fits a neural posterior
approximation to the exact likelihood of a single dataset, and a recent
flow-matching continuous normalizing-flow framework
\citep{liang2026accelerating} capable of accurately capturing the full
Hellings--Downs correlations in SGWB analyses.

Despite these advances, existing SBI efforts in PTA research have focused
almost exclusively on SGWB searches incorporating
per-pulsar noise models. The exploration of deterministic signals,
particularly individual inspiralling SMBHBs in eccentric orbits, remains
virtually absent in the SBI literature. At the same time, neural sequence
models have undergone rapid development. Transformers
\citep{Vaswani2017Attention}, originally introduced for natural language
processing tasks and popularized through large-scale language models~\citep{OpenAI_GPT6_2026}, have since been extended to computer
vision~\citep{Dosovitskiy:2020qjv} and sequence modeling applications
\citep{wen2022transformers}. More recently, Transformer architectures have begun
to appear in astronomy and cosmology
\citep{Hwang:2023oob,allam2024paying,Boersma:2024owk}, although their use
in GW data analysis remains less widespread than CNN- and RNN-based
approaches~\citep{cuoco2025applications}. However, Transformer-based modeling remains largely unexplored for PTA
time-series inference.

Transformers are particularly promising for PTA observations because
self-attention can capture long-range dependencies across an entire time
series, in contrast to recurrent architectures such as LSTMs~\citep{LSTM}
or convolution-based models that rely on sequential or local structure.
This is especially relevant for PTA residuals, where the signal from a
common GW source is weak, distributed across multiple pulsars, and
accumulates over long observing baselines. Recent work has further demonstrated the broader potential of Transformers
for GW inference across other frequency regimes. A pre-trained audio
Transformer such as OpenAI's Whisper can be fine-tuned to detect
LIGO/Virgo-band signals and classify noise transients
\citep{Chatterjee:2024ddw}. Recent Dingo-Pop work has also shown that
Transformer-based representations can enable end-to-end, amortized population
inference directly from GW strain data, while handling variable catalog
sizes~\citep{leyde2026end}. Dingo-T1 has demonstrated flexible parameter estimation for GW signals observed by ground-based detectors, with varying choices of detectors and frequency ranges~\citep{transgwkofler2026}. Vision Transformers have also been applied to glitch classification~\citep{srivastava2025vision}, while GraviBERT has shown that Transformer pretraining improves GW time-series inference and transfer across detector configurations and waveform models~\citep{benedikt2026gravibert}. However, purely data-driven models may still struggle to learn the complex
physics of orbital evolution and pulsar timing from finite training data.
As next-generation GW observatories produce larger datasets with increasingly
complex noise structure, physics-informed inductive biases become essential
for guiding representation learning toward physically meaningful and robust
inference~\citep{karniadakis2021physics}.

In this work, we introduce a novel approach that combines the representational power of transformers with a physics-informed positional encoding (PIPE), used alongside the standard sinusoidal encoding of vanilla transformers~\citep{Vaswani2017Attention}. In our framework, the transformer input is enriched with domain-specific quantities derived from the physical signal model, most notably the GW phase evolution of an SMBHB in a relativistic eccentric orbit. By embedding this information directly into the positional encoding, the network receives a sequence representation that is inherently aligned with the expected structure of PTA signals. This guides the model to attend to physically meaningful patterns, including orbital periodicity and pulsar-dependent timing contributions, while still allowing flexible data-driven learning.
\par Unlike traditional sinusoidal encodings, which can limit transformer performance when modeling complex GW signals, PIPE provides the model with physically structured coordinates that improve convergence, enhance feature extraction, and lead to more accurate parameter estimation. It is worth noting that PIPE-like physics-informed encodings have also been applied beyond GW astronomy, for example, to typhoon trajectory prediction from satellite imagery in climate science~\citep{li2025pipe}, highlighting their broader utility for integrating physical structure into deep learning architectures.

The remainder of this paper is organized as follows. Sec.~\ref{sec:gw_model} reviews the GW and PTA signal model for eccentric binaries, including the general relativistic post-Newtonian orbital dynamics used to generate simulated timing residuals. Sec.~\ref{sec:dataset} describes the dataset-generation procedure, including the construction of the synthetic multi-pulsar PTA dataset, the adopted noise model, and the realization-level signal-to-noise prescription. Sec.~\ref{sec:model-arch} introduces the proposed physics-informed simulation-based inference framework, including the Transformer backbone, phase-informed encodings, orbital-phase prediction module, and posterior density estimator. Sec.~\ref{sec:training} outlines the training strategy and implementation details, while Sec.~\ref{sec:evaluation} defines the evaluation metrics used to assess posterior accuracy, sharpness, and calibration. Sec.~\ref{sec:results} presents the main inference results, including comparisons between phase-informed and phase-agnostic models, posterior quality, and calibration performance. Finally, Sec.~\ref{sec:conclusion} summarizes the principal findings and discusses future directions toward more realistic and scalable PTA analyses.

\section{Gravitational waves and PTA signal model for eccentric binaries}
\label{sec:gw_model}
We begin by describing how GWs arise as solutions of the
Einstein field equations, the associated quadrupole formula, and the form of
GWs emitted by binaries in non-circular orbits in Subsection~\ref{sub:GW:gen}. This is
followed by a summary of the PTA signal produced by a supermassive black-hole
binary (SMBHB) inspiralling along a relativistic eccentric orbit. Computing the
resulting timing residuals requires modeling the binary's orbital dynamics
within the post-Newtonian (PN) framework, in which general-relativistic
corrections to Newtonian motion are systematically expanded in powers of
$(v/c)^2 \sim GM/(c^2 r)$,\footnote{A term of order $(v/c)^{2n}$ is referred to
as an $n$PN correction.} where $M$ is the total mass, $r$ the relative
separation, and $v$ the orbital velocity of the binary~\citep{Blanchet2014}.
Within this framework, the two GW polarization amplitudes $h_{+,\times}(t)$ can
be written explicitly in terms of the orbital variables, and the corresponding
PTA signal---the GW-induced timing residual---is obtained from their time
integrals.

The resulting expressions for the timing residual $R(t)$ are presented in
Subsection~\ref{sec:pta-signal-1}, while the PN-accurate orbital motion
underlying these expressions is detailed in
Subsection~\ref{sec:orbital-motion}. Alternative formulations for modeling PTA
signals from eccentric binaries may be found in
\citep{JenetLommen+2004,TaylorHuerta+2016,Susobhanan2023,SusobhananGopakumar+2020}.

\subsection{Generation of Gravitational Waves}
\label{sub:GW:gen}
GWs arise as radiative solutions of the Einstein field
equations,
\begin{equation}
G_{\mu\nu} = \frac{8\pi G}{c^4} T_{\mu\nu},
\end{equation}
where $G_{\mu\nu} \equiv R_{\mu\nu} - \tfrac12 g_{\mu\nu} R$ is the Einstein tensor,
$g_{\mu\nu}$ is the spacetime metric, $R_{\mu\nu}$ and $R$ are the Ricci tensor
and scalar, respectively, $T_{\mu\nu}$ is the stress--energy tensor of matter,
$G$ is Newton’s gravitational constant, and $c$ is the speed of light~\citep{thorne2000gravitation}.

In the weak-field regime relevant for GW observations, spacetime may be described
as a small perturbation $h_{\mu\nu}$ about flat spacetime,
\begin{equation}
g_{\mu\nu} = \eta_{\mu\nu} + h_{\mu\nu}, \qquad |h_{\mu\nu}| \ll 1 ,
\end{equation}
where $\eta_{\mu\nu}$ denotes the Minkowski metric. Linearizing the Einstein
equations to first order in $h_{\mu\nu}$ and imposing the Lorenz gauge condition
leads to wave equations for the metric perturbations, demonstrating that
gravitational disturbances propagate at the speed of light.

In vacuum, where $T_{\mu\nu}=0$, these equations admit propagating wave solutions.
Adopting the transverse--traceless (TT) gauge removes gauge redundancies and
isolates the two physical degrees of freedom of the gravitational field. For a
gravitational wave propagating in a given direction, the spatial components
$h_{ij}^{\rm TT}$ of the metric perturbation can be decomposed onto an
orthonormal polarization basis $\{e^{+}_{ij},\,e^{\times}_{ij}\}$ transverse to
the direction of propagation,
\begin{equation}
h_{ij}^{\rm TT}(t,\mathbf{x})
= h_{+}(t,\mathbf{x})\, e^{+}_{ij}
+ h_{\times}(t,\mathbf{x})\, e^{\times}_{ij},
\end{equation}
where $h_{+}$ and $h_{\times}$ denote the plus and cross polarization amplitudes.

At leading order in the weak-field, slow-motion expansion, GW emission from an isolated source is governed by the quadrupole formula~\citep{Einsteinquadrupole1918}
\begin{equation}
h^{\mathrm{TT}}_{ij}(t,\mathbf{x}) =
\frac{2G}{c^4 r}
\frac{d^2}{dt^2}
Q^{\mathrm{TT}}_{ij}(t-r/c),
\end{equation}
where $r$ is the distance to the source and $Q_{ij}$ is the mass quadrupole
moment of the source projected onto the TT basis.

For a compact binary system in a non-circular orbit, the resulting GW polarizations can be written
explicitly in terms of the orbital variables. At dominant quadrupole order, the plus and
cross polarizations are~\citep{gopakumar2002second}
\begin{align}
h^{Q}_{+}(r,\phi,\dot r,\dot\phi)
= -\frac{G M \eta}{D_L c^{4}}
\Bigg[
&(1+\cos^{2} i)
\Big(
\big(\tfrac{G M}{r} + r^{2}\dot\phi^{2} - \dot r^{2}\big)\cos 2\phi
+ 2 r \dot r \dot\phi \sin 2\phi
\Big)
\nonumber\\
&\quad + \sin^{2} i
\Big(
\tfrac{G M}{r} - r^{2}\dot\phi^{2} - \dot r^{2}
\Big)
\Bigg],
\end{align}
\begin{equation}
h^{Q}_{\times}(r,\phi,\dot r,\dot\phi)
= -\frac{G M \eta}{D_L c^{4}}
\, 2\cos i
\Big[
\big(\tfrac{G M}{r} + r^{2}\dot\phi^{2} - \dot r^{2}\big)\sin 2\phi
- 2 r \dot r \dot\phi \cos 2\phi
\Big],
\end{equation}
where $M=m_1+m_2$ is the total mass, $\eta=m_1 m_2/M^2$ is the symmetric mass
ratio, $r(t)$ and $\phi(t)$ are the orbital separation and phase,
$\dot r(t)$ and $\dot\phi(t)$ are their time derivatives, $i$ is the inclination
angle of the orbital angular momentum relative to the line of sight, and $D_L$
is the luminosity distance to the source.

\subsection{The PTA signal}
\label{sec:pta-signal-1}
Pulsar timing arrays probe these GW polarizations through their effect on pulse
times of arrival (TOAs). A passing GWs induce a timing residual
\begin{equation}
R(t)=\int_{t_0}^{t}\!\big[h(t')-h(t'-\Delta_p)\big]\,dt',
\end{equation}
where $\Delta_p = \frac{D_p}{c}(1-\cos\mu)$ encodes the pulsar distance $D_p$ and
the angle $\mu$ between the pulsar and the GW source~\citep{Detweiler1979}.  The
strain $h(t)$ combines the two polarization modes as
\begin{equation}
h(t)
= 
\begin{bmatrix} F_{+} & F_{\times} \end{bmatrix}
\mathbf{R}(2\psi)
\begin{bmatrix} h_{+}(t) \\ h_{\times}(t) \end{bmatrix},
\end{equation}
where $F_{+,\times}$ are the antenna pattern functions, $\psi$ is the GW
polarization angle, and $\mathbf{R}$ represents a rotation by $2\psi$.

Introducing $s_{+,\times}(t)=\int_{t_0}^t h_{+,\times}(t')\,dt'$, the residual may
be written as
\begin{align}\label{eq:residualterms}
R(t)=
\begin{bmatrix}F_{+} & F_{\times}\end{bmatrix}\mathbf{R}(2\psi)
\begin{bmatrix}
s_{+}(t)-s_{+}(t-\Delta_p) \\
s_{\times}(t)-s_{\times}(t-\Delta_p)
\end{bmatrix},
\end{align}
with the first and second terms in $s_{+,\times}$ corresponding to the Earth and pulsar
contributions.

For a SMBHB inspiralling along a relativistic eccentric
orbit, the orbital dynamics are modeled within the post-Newtonian (PN)
framework, in which corrections to Newtonian motion are systematically expanded
in powers of $(v/c)^2 \sim GM/(c^2 \, r)$~\citep{Blanchet2014}. Within this
framework, the orbital elements $r$, $\dot r$, $\phi$, and $\dot\phi$ can be expanded to the required PN order and used to construct the GW
polarizations $h_{+,\times} (t)$, which are subsequently time-integrated to obtain
$s_{+,\times} (t)$. However, the resulting exact PN expressions are generally
computationally expensive for large-scale data-analysis applications.
Under the assumption of slow periastron advance, the time-integrated GW
polarizations $s_{+,\times}$ admit closed-form analytic expressions that are
consistently expanded up to third post-Newtonian (3PN) order in the GW phase, as
derived in Refs.~\citep{JenetLommen+2004,Susobhanan2023}:

\begin{align}
s_{+}(t) &= \mathcal{S} \left[(1+c_\iota^2)(-\mathcal{P}\sin2\omega+\mathcal{Q}\cos2\omega)
 + s_\iota^2\,\mathcal{R}\right], \\
s_{\times}(t) &= 2 \, \mathcal{S} \,c_\iota\left(\mathcal{P}\cos2\omega+\mathcal{Q}\sin2\omega\right),
\end{align}
where $c_\iota=\cos\iota$, $s_\iota=\sin\iota$, and the functions
\begin{subequations}
\begin{align}
\mathcal{P} &=\frac{\sqrt{1-e_t^{2}}(\cos2u-e_t\cos u)}{1-e_t\cos u}, \\
\mathcal{Q} &=\frac{[(e_t^{2}-2)\cos u + e_t]\sin u}{1-e_t\cos u}, \\
\mathcal{R} &= e_t\sin u,
\end{align} 
\end{subequations}
depend on the time eccentricity $e_t$ and eccentric anomaly $u$.

The amplitude factor is $\mathcal{S}=\mathcal{H}/n$, with
$\mathcal{H} = (GM\eta/D_L c^2)\,x$, where $x=[GM(1+k)n/c^3]^{2/3}$ is the
dimensionless PN parameter, $n$ is the mean orbital motion, $k$ is the
relativistic periapsis advance per orbit, $\omega=\phi-f$ is the argument of
periapsis, and $\phi$ is the orbital phase. The chirp mass is
$\mathcal{M}_c=\eta^{3/5}M$.

\subsection{PN-accurate orbital motion}
\label{sec:orbital-motion}
Up to this point, the PN formalism has been considered only in the conservative
evolution, which accounts for periastron advance without considering the
reactive evolution, i.e.\ the GW-driven decay of the orbital elements.  
To compute $R(t)$, we also require the evolution of the orbital parameters.  
The dynamics separate into conservative evolution (periapsis advance) and
reactive evolution (GW-driven decay). For the conservative part, we model  dynamics of the
relativistic eccentric binary using the PN-accurate quasi-Keplerian
parametrization \citep{DamourDeruelle1985,MemmesheimerGopakumarSchafer2004}. We adopt the PN quasi-Keplerian parametrization with mean anomaly
\begin{equation}
l(t)=\int_{t_0}^t n(t')\,dt'
\end{equation}
and the PN-accurate Kepler equation~\citep{MemmesheimerGopakumarSchafer2004,BoetzelSusobhanan+2017}
\[
l = u - e_t\sin u + \mathfrak{F}_t(u),
\]
where $\mathfrak{F}_t$ is periodic in $u$.  
The orbital GW phase is
\begin{equation}
\label{GW:ph}
\phi = \gamma + l + (1+k)(f-l) + \mathfrak{F}_\phi(u) \, ,    
\end{equation}
with true anomaly
\[
f=2\arctan\!\left[\sqrt{\frac{1+e_\phi}{1-e_\phi}}\,\tan\frac{u}{2}\right].
\]
The argument of periastron is $\omega=\phi-f$, while the periastron angle evolves as  
\[
\gamma(t)=\int_{t_0}^t k(t')n(t')\,dt'.
\]

The GW-driven evolution of the orbital elements is governed by the following 2.5PN accurate (absolute)
radiation–reaction equations \citep{DamourGopakumarIyer2004}:
\begin{subequations}
\begin{align}
\frac{dn}{dt} &= \frac{1}{5}\!\left(\frac{GMn}{c^{3}}\right)^{5/3}\eta \, n^{2}\frac{96+292e_{t}^{2}+37e_{t}^{4}}{(1-e_{t}^{2})^{7/2}}, \\
\frac{de_{t}}{dt} &= -\frac{1}{15}\!\left(\frac{GMn}{c^{3}}\right)^{5/3}\eta \, n \, e_{t}\frac{304+121e_{t}^{2}}{(1-e_{t}^{2})^{5/2}}, \\
\frac{d\gamma}{dt} &= k \, n, \\
\frac{dl}{dt} &=n.
\end{align}
\end{subequations}
Solving these coupled equations yields $n(t)$, $e_t(t)$, $\gamma(t)$, and $l(t)$; combining them with the equations for $u$, $f$, and $\omega$ provides all ingredients needed to evaluate the PTA signal $R(t)$. 
\par
\begin{figure*}[t]
\centering
\includegraphics[width=\textwidth]{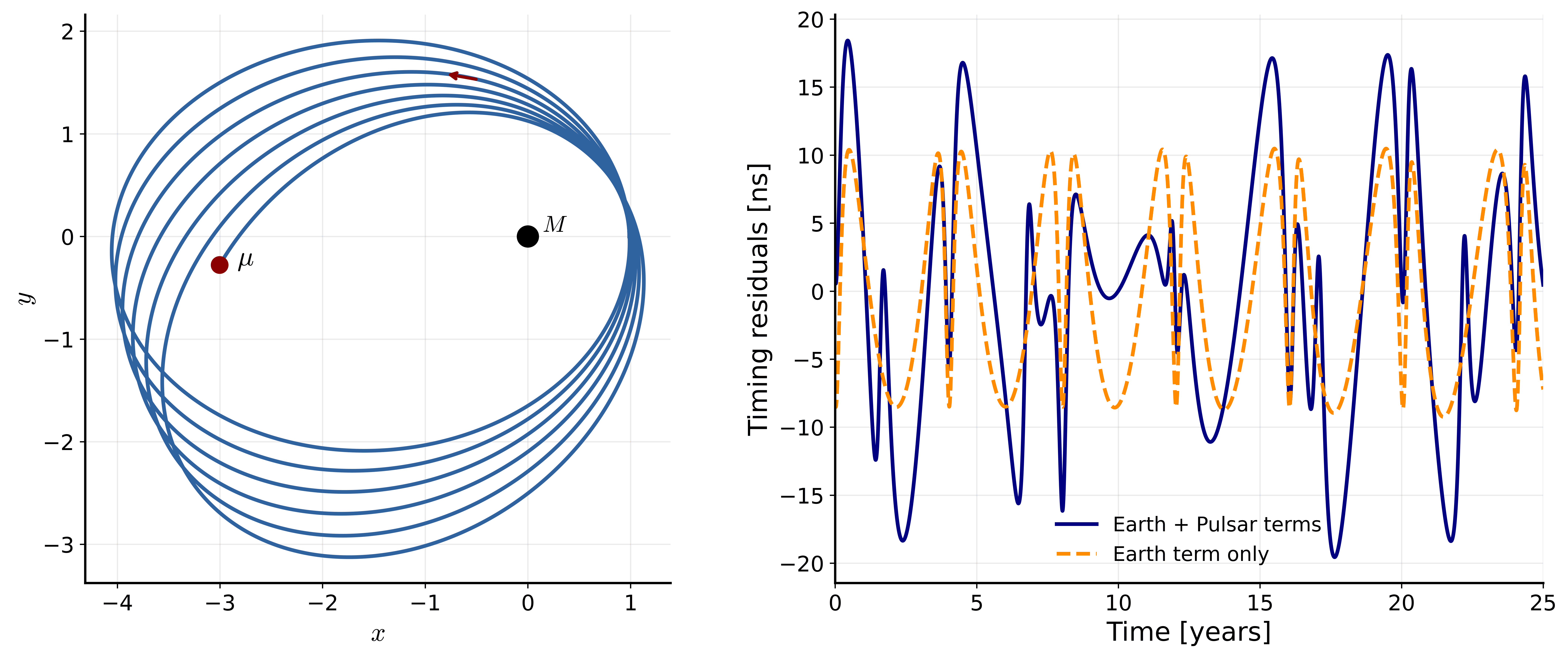}
\caption{
Left: Orbital trajectory of the secondary black hole about the primary
(fixed at the origin) in the center-of-mass frame.
Relativistic periastron precession due to post-Newtonian effects is visible.
Spatial coordinates are shown in units of $100\,GM/c^2$.
The arrow indicates the direction of motion.
Parameters used are
$M=10^{9}M_\odot$,
$q=1$,
$P=4\,\mathrm{yr}$,
$e_0=0.6$,
$\iota=0$,
$\psi=0$,
$\gamma_0=0$,
$l_0=0$,
$D_{\mathrm{GW}}=1\,\mathrm{Gpc}$,
and $t_{\mathrm{ref}}=0$.
Right: Corresponding induced PTA residuals for PSR~J1909$-$3744
over $t\in[0,25]\,\mathrm{yr}$,
showing the full response (the Earth + pulsar terms in Eq.~\eqref{eq:residualterms}) and the response including only the Earth term.
}
\label{fig:orbit_pta_residuals}
\end{figure*}

 In our analysis, the initial time--eccentricity $e_t$ is denoted by $e_0$, a
convention we adopt throughout. Likewise, we use $n_0$ to denote the initial
mean motion, as summarized in Table~\ref{tab:pulsars}. Unless otherwise
specified, the initial values of the periastron angle $\gamma$ and the mean
anomaly $l$ are fixed to zero for all simulations. 
As a representative visual demonstration, Fig.~\ref{fig:orbit_pta_residuals}
shows the fully 3PN eccentric orbital evolution and the corresponding PTA
timing-residual response generated by the signal model.

\section{Dataset generation}
\label{sec:dataset}

We construct a synthetic PTA dataset tailored for supervised learning of
continuous-wave binary parameters. Our simulations focus on ten well--timed PTA
pulsars that also contribute prominently to the stochastic gravitational-wave
background (SGWB) evidence in the NANOGrav 15-year dataset~\citep{agazie2023ng15yrgwb}.
Their distances are summarized in Table~\ref{tab:pulsars}, and they form the
basis for all synthetic timing-residual realizations used in this study.

\begin{table}[!h]
\centering
\caption{The ten PTA pulsars used in this study, with their distances
$p_{\rm dist}$ in kpc.}
\label{tab:pulsars}
\begin{tabular}{lclc}
\hline\hline
Pulsar & $p_{\rm dist}$ [kpc] & Pulsar & $p_{\rm dist}$ [kpc] \\
\hline
J1909$-$3744 & 1.26 & J0030+0451   & 0.28 \\
J2317+1439   & 1.89 & J1910+1256   & 1.95 \\
J2043+1711   & 1.00 & J1744$-$1134 & 0.42 \\
J1600$-$3053 & 2.40 & J1944+0907   & 1.00 \\
J1918$-$0612 & 1.40 & J0613$-$0200 & 0.90 \\
\hline\hline
\end{tabular}
\end{table}

For each pulsar $j$, we simulate the timing-residual time series
\begin{equation}
    d_{ji} = s_{ji} + n_{ji},
\end{equation}
where $i$ indexes the discrete observation times $t_i$,
$s_{ji} = s(t_i, \boldsymbol{\lambda})$ denotes the deterministic
GW-induced residual for source parameters $\boldsymbol{\lambda}$, and $n_{ji}$
represents additive white Gaussian noise with rms $\sigma$. Each realization
yields a residual vector $\mathbf{d}_j = (d_{j1}, \dots, d_{jL})$ of fixed length
$L = 400$.

Unless otherwise stated, we generate $5\times10^4$ independent realizations per pulsar by drawing the binary and geometric parameters from Table~\ref{tab:dataset-summary}, yielding $5\times10^5$ simulated time series across the 10 pulsars. This dataset forms the default training and validation set used for the posterior-inference experiments in this work. For two specific studies, we additionally employ an expanded dataset comprising $4\times10^5$ realizations: first, for the realization-level phase (and signal-to-noise ratio) prediction analysis presented in Sec.~\ref{result:ph:pred}, and second, for posterior comparison with conventional Bayesian inference in Sec.~\ref{result:large:data}. The supervised learning targets are the intrinsic parameter quartet
$\boldsymbol{\theta}\equiv(n,e_0,M,\mathcal{S})$, which characterizes the eccentric binary black hole (EBBH) system. The same framework readily extends to
higher-dimensional inference, and in Sec.~\ref{result:large:data} we demonstrate
a realistic Bayesian posterior comparison over the higher-dimensional EBBH parameter space. Prior to training, all inputs and targets are normalized according
to
\begin{equation*}
x \;\mapsto\; \frac{x - \mu_{\rm train}}{\sigma_{\rm train}},
\end{equation*}
where $\mu_{\rm train}$ and $\sigma_{\rm train}$ denote the mean and standard
deviation computed exclusively from the training subset.

The simulated dataset is prepared in realization mode, where all $P=10$
pulsars associated with a given GW source realization are retained jointly.
Each training sample is represented as a tensor
\begin{equation}
\mathbf{X} = \big[\, \mathbf{X}_1,\dots,\mathbf{X}_P \,\big],
\end{equation}
corresponding to the stacked multi-pulsar timing-residual time series for a
single realization, where $\mathbf{X}_j \in \mathbb{R}^{L}$ denotes the residuals
for pulsar $j$. This representation is natural for PTA inference, as
information about a common GW source is distributed across the array and is
most appropriately characterized at the realization level.

\begin{table}[t]
\centering
\caption{Summary of the simulated PTA dataset used for training and validation.}
\label{tab:dataset-summary}
\renewcommand{\arraystretch}{1.2}
\begin{tabular}{l l}
\hline\hline
Realizations per pulsar & $5\times10^4$ samples \\[2pt]

\multirow{4}{*}{Sampled parameters}
& $\log_{10} n \in [-8,\,-6.7]~{\rm Hz}$,\quad $e_0 \in [0.1,\,0.8]$ \\
& $\log_{10} M \in [7,\,10]~M_\odot$,\quad $\log_{10}\mathcal{S} \in [-8,\,-6]~{\rm s}$ \\
& $\cos\theta \in [0,1]$,\quad $\phi \in [0,\,2\pi]~{\rm rad}$ \\
& $q \in [0.1,\,1]$,\quad $\cos\iota \in [0,\,1]$,\quad $\psi \in [0,\,\pi]~\mathrm{rad}$ \\[3pt]

Noise model & White Gaussian, $C = \sigma^2 I$ \\
Targets & $(n_0, e_0, M, \mathcal{S})$ (extendable) \\
\hline\hline
\end{tabular}
\end{table}

To emulate realistic PTA-like noise conditions, we assign each realization a
target signal-to-noise ratio (SNR) using the standard PTA covariance-weighted
definition~\citep{petrov2024identifying,taylor2021nanohertz}
\begin{equation}
\mathrm{SNR}^2 \equiv \mathbf{s}^{\top}\mathbf{C}^{-1}\mathbf{s},
\end{equation}
where $\mathbf{s} = [\,\mathbf{s}_1,\dots,\mathbf{s}_P\,]$ is the stacked noiseless multi-pulsar
signal and $\mathbf{C}$ is the full PTA noise covariance matrix. In general, $\mathbf{C}$ encodes~\citep{van2013understanding} contributions from intrinsic pulsar red noise, dispersion-measure (DM) variations, and inter-pulsar correlations such as the Hellings--Downs (HD) spatial correlations induced by a common gravitational-wave background, such that detectability is governed by this global noise-weighted quadratic form. In this work, we adopt a simplified white-noise (WN) model, for which these effects are neglected and the covariance reduces to a diagonal form. 

Equivalently, defining per-pulsar contributions,
$\mathrm{SNR}_j^2 = (s_j \mid s_j)$, the combined PTA SNR is
$\mathrm{SNR}_{\mathrm{com}} = \sqrt{\sum_{j=1}^{P} \mathrm{SNR}_j^2}$,
which satisfies
$\mathrm{SNR}_{\mathrm{com}}^2 = \mathbf{s}^{\top}\mathbf{C}^{-1}\mathbf{s}$
for block-diagonal covariance. In this work, we therefore identify the
realization-level SNR as
$\mathrm{SNR}_r \equiv \mathrm{SNR}_{\mathrm{com}}$.

For the white-noise-only model adopted here, the covariance reduces to
$\mathbf{C} = \sigma_r^2 \, \mathbf{I}$, so that
\begin{equation}
\label{eq:snr_realisation_wn}
\mathrm{SNR}_r^2 = \frac{1}{\sigma_r^2}\|\mathbf{s}_r\|_2^2.
\end{equation}
Accordingly, for each realization $r$, we draw
$\mathrm{SNR}_r \sim \mathsf{Uniform}(\mathrm{SNR}_{\min},\,\mathrm{SNR}_{\max})$,
and set the corresponding noise amplitude as
\begin{equation}
\sigma_r = \frac{\|\mathbf{s}_r\|_2}{\mathrm{SNR}_r}.
\end{equation}
Independent Gaussian noise samples $n_{ji}\sim\mathcal{N}(0,\sigma_r^2)$ are
then added to produce the noisy PTA realization
\begin{equation}
X^{\mathrm{noisy}}_{ji} = s_{ji} + n_{ji},
\end{equation}
ensuring that each realization attains the prescribed composite (array-level)
SNR by construction.

The same dataset is subsequently used to train and evaluate the posterior-inference pipeline, where a physics-informed Transformer provides learned representations for simulation-based inference. This ensures a unified and internally consistent training--evaluation setup. The model architecture and inference components are described in the following sections, Secs.~\ref{subsec:backbone} and~\ref{subsec:flows}.

\section{Model Architecture}
\label{sec:model-arch}

We now describe the physics-informed simulation-based inference architecture
used in this work. Each input sample is a full PTA realisation
$\mathbf{X}\in\mathbb{R}^{P\times L}$, containing $P$ pulsars and $L$ time
samples per pulsar. The model maps this multi-pulsar residual tensor to a
compact context vector that conditions a normalizing-flow posterior
estimator.

The pipeline consists of two coupled stages, summarized in
Fig.~\ref{fig:model-architecture}. First, a hierarchical Transformer extracts temporal structure within each pulsar and cross-pulsar
correlations across the PTA, using standard positional encodings together
with orbital-phase physics-informed encodings. Second, the resulting context vector conditions a normalizing-flow density
estimator for the posterior $p(\boldsymbol{\theta}\,|\,\mathbf{X})$, enabling
amortized posterior sampling while quantifying uncertainties, correlations,
and degeneracies in the inferred EBBH parameters.

\subsection{Physics-Informed Transformer Backbone}
\label{subsec:backbone}

The fundamental operation underlying transformers is the attention mechanism, which learns token–token relationships using trainable query ($Q$), key ($K$), and value ($V$) projections. Given an input token representation $Z$, the query, key, and value matrices are obtained through learned linear projections, and the self-attention representation is computed as~\citep{Vaswani2017Attention}
\begin{equation}
\mathrm{Attention_{self}}=\mathsf{softmax} \ \!\big(QK^{\top}/\sqrt{d_h}\big)V,
\end{equation}
with $d_h$ the head dimension. While multi-head attention captures a rich set of token interactions, self-attention alone only encodes relationships within each individual sample.  


\subsubsection{Physics-Informed Encodings:}
\label{subsec:phys-enc}
Standard transformer encodings rely solely on token indices, which are agnostic to the physics of GW signals and the heterogeneous nature of PTA datasets. To address this limitation, we incorporate physics-informed embeddings, specifically orbital-phase encoding, in addition to the standard
sinusoidal positional encoding. This ensures that each token is enriched with astrophysically meaningful structure before being processed by the Transformer encoder.
\paragraph{Sinusoidal positional encoding:} We retain the canonical sinusoidal positional encoding to endow the model with an explicit sense of sequential ordering~\citep{Vaswani2017Attention}:
\begin{equation}
\mathrm{PE}^{pos}(s,2i)=\sin\!\left(\frac{s}{10000^{2i/d}}\right),\qquad
\mathrm{PE}^{pos}(s,2i{+}1)=\cos\!\left(\frac{s}{10000^{2i/d}}\right),
\end{equation}
where $s$ indexes the token position and $d$ is the model dimension. A learnable gate $\omega_{\mathrm{pos}}$ modulates its influence.

\paragraph{Physics-informed orbital-phase encoding:} Eccentric binary black holes exhibit a temporal evolution dominated by the orbital phase $\phi(t)$ as described in Eq.~\eqref{GW:ph}, which encodes periastron advance and waveform modulation.  
Index-based encodings cannot capture this physical structure, especially under irregular sampling or low signal-to-noise conditions.  

To embed domain-specific information, we construct a physics-informed phase encoding (PIPE).  
The instantaneous orbital phase is computed for each waveform and wrapped into $(-\pi,\pi]$. Further, $\phi(t)$ is averaged within each patch to obtain $\phi^{tok}$.  
This is mapped into a sinusoidal basis:
\begin{equation}
\label{eq:phase-encoding}
\mathrm{PE}^{\phi}(b,s,2i)=\sin\!\left(\frac{\phi^{tok}_{b,s}}{10000^{2i/d}}\right),\qquad
\mathrm{PE}^{\phi}(b,s,2i{+}1)=\cos\!\left(\frac{\phi^{tok}_{b,s}}{10000^{2i/d}}\right).
\end{equation}
This yields a phase-informed embedding tensor 
$\mathrm{PE}^{\phi}\in\mathbb{R}^{B\times S\times d}$, 
where $B$ is the batch size, $S$ the number of tokens, 
and $d$ the embedding dimension. A learnable gate $\omega_{\phi}$ calibrates the relative strength of this physics-based contribution.

\paragraph{Combined positional–phase representation:}

In practice, the model learns how strongly to rely on positional and
physics-informed structure through two trainable scalar gates,
$\omega_{\mathrm{pos}}$ and $\omega_{\phi}$.  The fused token embedding is
written compactly as
\begin{equation}
\label{eq:Z:T}
z_s = T_s + \sum_{k\in\{\mathrm{pos},\phi\}} \omega_k\,\mathrm{PE}^{(k)}_s ,
\end{equation}
where the weights $\omega_k$ determine the relative influence of the positional
encoding and the orbital-phase encoding. These gates are optimized jointly with
all other parameters. Their gradients take the form
\begin{equation}
\frac{\partial \mathcal{L}}{\partial \omega_k}
=
\sum_{s}
\Big\langle
\frac{\partial \mathcal{L}}{\partial z_s},\;
\mathrm{PE}^{(k)}_s
\Big\rangle,
\qquad k\in\{\mathrm{pos},\phi\},
\label{eq:encoding-weight-gradient}
\end{equation}
where $\langle \cdot,\cdot \rangle$ denotes the standard Euclidean inner product
in the embedding space, i.e.\ $\langle a,b\rangle = a^\top b$.
Consequently, each gate increases when its encoding aligns with the
backpropagated error signal and decreases otherwise. This allows the model to adjust
automatically how much it relies on physical or positional structure, depending
on the information content and noise level of the PTA residuals.

\subsubsection{Patch Embedding, Hierarchical Transformer, and Point-Estimate Head:}
\label{subsec:ea-head}

\par Following the construction of physics-informed token embeddings, each PTA
realisation is processed by a hierarchical Transformer architecture designed to
separate intra-pulsar temporal modeling from inter-pulsar information exchange.
Each input sample corresponds to a full PTA realisation,
$\mathbf{X}=[\mathbf{X}_1,\dots,\mathbf{X}_P]$, where
$\mathbf{X}_j\in\mathbb{R}^{L}$ denotes the timing-residual sequence of pulsar
$j$. This joint representation preserves the array-level structure of the data,
allowing the model to learn both temporal features within individual pulsars and
correlations across the pulsar ensemble.

For each pulsar $j$, the sequence is divided into
$S=L/P_{\mathrm{patch}}$ non-overlapping temporal patches of size
$P_{\mathrm{patch}}$,
$\mathbf{X}_j\rightarrow\{\mathbf{x}_{j,s}\}_{s=1}^{S}$, with
$\mathbf{x}_{j,s}\in\mathbb{R}^{P_{\mathrm{patch}}}$~\citep{Dosovitskiy:2020qjv}. Each patch therefore
acts as a localized temporal token, reducing the effective sequence length from
$L$ to $S$ while retaining short-range correlations within each segment. These patches are mapped into the latent space using a shared linear projection followed by a Gaussian Error Linear Unit (GELU) activation~\citep{hendrycks2016gaussian},
\begin{equation}
\mathbf{t}_{j,s}=\mathsf{GELU}(W_e\mathbf{x}_{j,s}+b_e),
\qquad
W_e\in\mathbb{R}^{d\times P_{\mathrm{patch}}},
\end{equation}
where $\mathsf{GELU}(x)=x\,\Phi(x)$ with $\Phi(x)$ the standard normal CDF, which smoothly gates activations and has been shown to outperform ReLU~\citep{nair2010rectified} in Transformer-based architectures. This projection produces the embedded token sequence $T_j=[\mathbf{t}_{j,1},\dots,\mathbf{t}_{j,S}]\in\mathbb{R}^{S\times d}$.
Thus, patching defines the temporal support of each token, while the embedding
projection defines its representation in feature space. Standard positional encodings and, when enabled, physics-informed phase
encodings, as defined in Eq.~\eqref{eq:phase-encoding}, are then added to
obtain the initial token representation $Z_j^{(0)}$. As an optional alternative
to explicit patch projection, a lightweight convolutional front-end
(Conv--BN--GELU--pool) may be used to extract short-range temporal structure
before token formation, yielding $T_{\mathrm{stem}}\in\mathbb{R}^{S\times d}$.

Each pulsar token sequence $Z_j^{(0)}$ is then processed independently by a
stack of $D$ multi-head Transformer blocks~\citep{Vaswani2017Attention}, yielding
$Z_j^{\mathrm{temp}}=\mathcal{F}_{\mathrm{T}}\left(Z_j^{(0)}\right)$.
For each block with $h$ heads and head dimension $d_h=d/h$, the standard
linear projections generate
$Q,K,V\in\mathbb{R}^{h\times S\times d_h}$, where $S$ is the number of temporal
tokens. The attention weights and contextual features are computed as
\begin{equation}
A=\mathrm{Softmax}\left(
\frac{QK^\top}{\sqrt{d_h}}
\right),
\qquad
C=\mathrm{Concat}\left(A V\right).
\end{equation}
This operation allows each temporal token to aggregate information from the
full pulsar sequence, enabling the encoder to capture long-range temporal
dependencies in the timing residuals. The temporal encoder output is summarized for each pulsar by mean
pooling over tokens,
\begin{equation}
\mathbf{z}_j=\frac{1}{S}\sum_{s=1}^{S}Z_{j,s}^{\mathrm{temp}}
\in\mathbb{R}^{d},
\end{equation}
yielding a single latent embedding for each pulsar.

To model array-level correlations, the pulsar embeddings are treated as a
sequence
$Z^{\mathrm{psr}}=[\mathbf{z}_1,\dots,\mathbf{z}_P]\in\mathbb{R}^{P\times d}$
and passed through a cross-pulsar attention module,
\begin{equation}
\mathrm{Attn}_{\mathrm{psr}}=
\mathrm{Softmax}\!\left(
\frac{Q_{\mathrm{psr}}K_{\mathrm{psr}}^\top}{\sqrt{d_h}}
\right)V_{\mathrm{psr}} .
\end{equation}
This step enables information exchange across pulsars, allowing each pulsar representation to incorporate global array-level information and capture correlations induced by common GW signals and, where applicable, structured noise (such as correlated pulsar noise with Hellings--Downs correlations~\citep{hellings1983upper}).

Finally, a permutation-invariant aggregation produces the realisation-level
context vector
\begin{equation}
\label{h:context}
h=\frac{1}{P}\sum_{j=1}^{P}\mathrm{Attn}_{\mathrm{psr},j}.
\end{equation}
The context vector $h$ summarizes both the intra-pulsar temporal structure and
the inter-pulsar correlations, and serves as the conditioning input to the
posterior density estimator described in Sec.~\ref{subsec:flows}. In addition,
a lightweight regression head can also be used to produce point estimates,
$\hat{y}=f_{\mathrm{MLP}}(h)$, for selected physical parameters. Although not
used in the posterior-inference results presented here, this head provides an
optional auxiliary supervised signal and an interpretable summary of the latent
state. Overall, the hierarchical design provides a scalable,
covariance-aware approximation to PTA inference by explicitly separating
temporal modeling within pulsars from relational modeling across the pulsar
array.

\begin{figure*}[t]
\centering
\includegraphics[width=\textwidth]{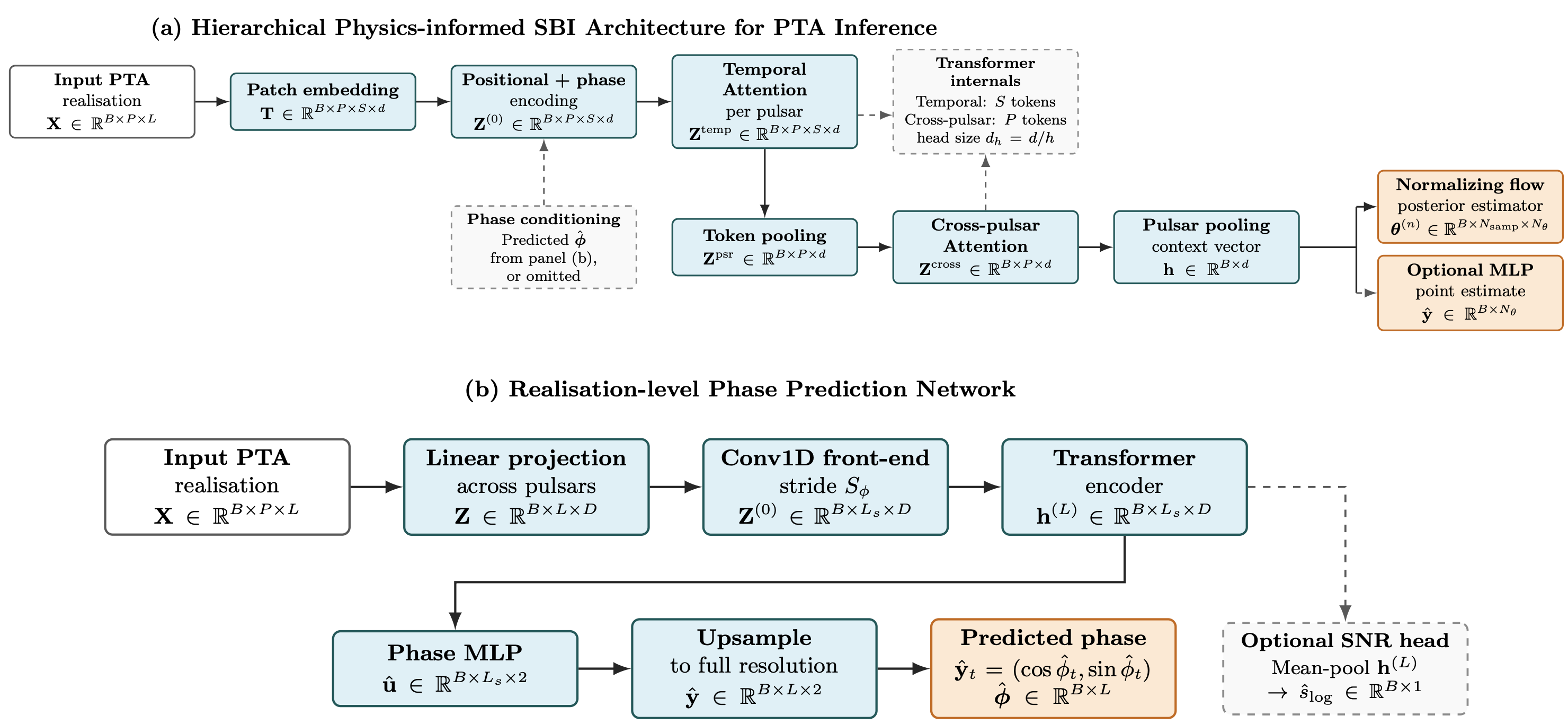}
\caption{
Combined overview of the phase-informed simulation-based inference pipeline.
\textbf{(a)} The hierarchical physics-informed SBI architecture uses patch embedding,
positional and optional phase encoding, temporal attention blocks, token pooling,
and cross-pulsar attention to form a global context vector $\mathbf{h}$.
This context conditions a normalizing-flow posterior estimator for
$\boldsymbol{\theta}$, while the dashed branch denotes an optional point-estimate head.
\textbf{(b)} The realisation-level phase prediction network maps the noisy multi-pulsar PTA input
to a shared predicted orbital-phase trajectory $\hat{\boldsymbol{\phi}}$ and, optionally, a
realisation-level SNR estimate. The predicted phase is then used as the optional
phase-conditioning input in panel (a).
Here, $B$ is the batch size, $P$ the number of pulsars, $L$ the number of time samples,
$S$ is the number of temporal patches in panel (a), $S_\phi$ is the temporal stride in panel (b),
$L_s=L/S_\phi$ is the downsampled phase-model sequence length, $d$ and $D$ are embedding
dimensions, $h$ is the number of attention heads, $d_h=d/h$, $N_\theta$ is the number of inferred
parameters, and $N_{\rm samp}$ is the number of posterior samples.
}
\label{fig:combined_phase_sbi_architecture}
\end{figure*}

\subsection{Phase Prediction Network}
\label{subsec:phase_net}

The physics-informed encoding requires access to the instantaneous orbital phase $\phi(t)$ given by Eq.~\eqref{GW:ph}, which is not directly observable in realistic PTA datasets. Instead, the observed data consist of noisy timing residuals from multiple pulsars, all of which share a common underlying phase evolution. Therefore, phase inference is formulated as a realisation-level task, where the model jointly processes all pulsars to recover a single global phase trajectory.

For each realisation, the input is $X \in \mathbb{R}^{P \times L}$ with $X_{a,t} = x_a(t)$, where $P$ is the number of pulsars and $L$ is the number of time steps. The shared phase trajectory is represented on the unit circle as $y_t = (\cos\phi_t, \sin\phi_t)$.

At each time step, the $P$-dimensional multi-pulsar residual vector is projected into a latent embedding using a learnable linear map $W_{\mathrm{in}} \in \mathbb{R}^{D \times P}$,
\begin{equation}
z_t = W_{\mathrm{in}} X_{:,t}, \qquad z_t \in \mathbb{R}^{D},
\end{equation}
forming a sequence $Z \in \mathbb{R}^{L \times D}$. This projection aggregates information across pulsars and acts as an implicit denoising mechanism.

For computational efficiency, we adopt a ``fast'' Transformer architecture with a convolutional front-end. The embedded sequence is processed by a 1-D convolutional network with stride $S$ along the temporal dimension, reducing the sequence length to $L_s = L/S$. After layer normalization and addition of sinusoidal positional encoding $p_i$, we obtain $h_i^{(0)} = \mathrm{LayerNorm}(z_i^{(0)}) + p_i$ for $i=1,\dots,L_s$.

The sequence is then processed by a stack of Transformer encoder layers with multi-head self-attention and position-wise feedforward networks,
\begin{equation}
h^{(L)} = \mathrm{TransformerEncoder}(h^{(0)}), \qquad h^{(L)} \in \mathbb{R}^{L_s \times D}.
\end{equation}

A token-wise multilayer perceptron maps the latent features to low-rate phase vectors $u_i = \mathrm{MLP}_{\phi}(h_i^{(L)}) \in \mathbb{R}^2$, which are upsampled back to the original temporal resolution to obtain $\hat{y}_t$, interpreted as $(\cos\hat{\phi}_t, \sin\hat{\phi}_t)$.

Optionally, a global representation $\bar{h} = \frac{1}{L_s} \sum_i h_i^{(L)}$ can be used to regress a scalar SNR, defined at the realisation level as in Eq.~\eqref{eq:snr_realisation_wn}, via a lightweight MLP, enabling joint learning of phase and global signal strength from multi-pulsar inputs.

\subsubsection{Multi-Component Phase and SNR Loss:}
\label{subsec:losses}

The training objective combines complementary losses defined on the unit-circle representation of phase, along with an SNR regression term.

Let $\hat{y}_j, y_j \in \mathbb{R}^2$ denote predicted and true phase vectors, with normalized versions $\tilde{y}_j$ and $\tilde{t}_j$. The predicted phase angle is
\begin{equation}
\hat{\phi}_j = \operatorname{atan2}(\hat{y}_{j,2}, \hat{y}_{j,1}).
\end{equation}

\paragraph{Von Mises alignment loss:}
This term enforces agreement between predicted and true phase on the unit circle.
\begin{equation}
L_{\mathrm{von}} = - \, \mathbb{E}_j[\tilde{y}_j^\top \tilde{t}_j].
\end{equation}

\paragraph{Circular smoothness loss:}
This term promotes temporally smooth phase evolution by penalizing rapid phase changes.
\begin{equation}
d_j = \mathrm{wrap}_{[-\pi,\pi)}(\hat{\phi}_{j+1}-\hat{\phi}_j), \qquad
L_{\mathrm{smooth}} = \mathbb{E}_j[d_j^2].
\end{equation}

\paragraph{Spectral consistency loss:}
This term ensures that the predicted phase trajectory matches the global frequency structure of the true signal.
We form complex sequences $z_j = \cos\phi_j + i\sin\phi_j$ and $\hat{z}_j = \cos\hat{\phi}_j + i\sin\hat{\phi}_j$, compute their Fourier transforms $T = \mathrm{FFT}(z)$ and $P = \mathrm{FFT}(\hat{z})$, and compare normalized magnitudes:
\begin{equation}
\tilde{T} = \frac{|T|}{\|\,|T|\,\|_2}, \qquad
\tilde{P} = \frac{|P|}{\|\,|P|\,\|_2}, \qquad
L_{\mathrm{spec}} = \|\tilde{P} - \tilde{T}\|_2^2.
\end{equation}

The log-SNR $s_{\log} = \log_{10} s$ is learned using a mean-squared error, capturing the global signal strength of the realisation,
\begin{equation}
L_{\mathrm{SNR}} = (\hat{s}_{\log} - s_{\log})^2.
\end{equation}

The overall loss is given by
\begin{equation}
L_{\mathrm{total}} =
\kappa_{\mathrm{von}}  L_{\mathrm{von}} 
+ \alpha_{\mathrm{smooth}} L_{\mathrm{smooth}}
+ \alpha_{\mathrm{spec}} L_{\mathrm{spec}}
+ \lambda_{\mathrm{SNR}} L_{\mathrm{SNR}}.
\end{equation}
Here, $\kappa_{\mathrm{von}}, \alpha_{\mathrm{smooth}}$, $\alpha_{\mathrm{spec}}$, and $\lambda_{\mathrm{SNR}}$ are weighting coefficients that control the relative contributions of the smoothness, spectral-consistency, and optional realisation-level SNR-estimation terms, respectively. Overall, this objective enforces phase alignment, temporal smoothness, spectral consistency, and, when included, accurate realisation-level SNR estimation. Throughout this work, we use fixed loss weights $\kappa_{\mathrm{von}}=8$, $\alpha_{\mathrm{smooth}}=0.10$, $\alpha_{\mathrm{spec}}=0.05$, and $\lambda_{\mathrm{SNR}}=0.05$. These coefficients were chosen empirically to balance the contributions of the individual loss terms while maintaining the von Mises objective as the primary optimization target.
\subsection{Posterior Density Estimation via Normalizing Flows}
\label{subsec:flows}
The hierarchical Transformer backbone maps each PTA realisation to a physics-informed
context vector $\mathbf{h}$ that summarizes temporal structure within pulsars
and correlations across the array. For scientific inference, we require the
full conditional posterior distribution
$p(\boldsymbol{\theta}\,|\,\mathbf{x})$ in order to quantify uncertainties,
parameter correlations, degeneracies, and possible multi-modality. We therefore
condition a normalizing-flow density estimator on $\mathbf{h}$ to perform
amortized posterior inference.

The flow learns an invertible transformation between a simple latent Gaussian
and the target posterior over EBBH parameters, conditioned on the encoded PTA
time-series representation (see Part~2 of Fig.~\ref{fig:model-architecture}).
In this work, we primarily use a \emph{Discrete Normalizing Flow} (DNF)~\cite{Rezende2015,Papamakarios2019} based
on affine coupling layers, with a \emph{Continuous Normalizing Flow} (CNF)~\citep{Chen2018} included as an auxiliary comparison. Both models share the same conditioning
backbone and differ only in the invertible transformation used to represent the
conditional density.

\paragraph{Conditioning network for the flow:}
Both DNFs and CNFs condition on the same learned representation $\mathbf{h}$, as defined in Eq.~\eqref{h:context}. For completeness, the conditioning path can also be viewed as operating directly on the input $\mathbf{x} \in \mathbb{R}^{L}$ through the same convolutional–Transformer encoder, phase-informed embeddings, and pooling steps, ultimately producing $\mathbf{h} \in \mathbb{R}^{B \times d}$ as specified in Eq.~\eqref{h:context}.

\subsubsection{Discrete Normalizing Flow (DNF):}
\label{subsubsec:dnf}

The DNF models the conditional posterior $p(\boldsymbol{\theta}\,|\,\mathbf{x})
\equiv p(\boldsymbol{\theta}\,|\,\mathbf{h})$
via a sequence of $K$ invertible transformations applied to a simple latent 
variable~\cite{Rezende2015,Papamakarios2019}.  
Let $\mathbf{z} \sim p_0(\mathbf{z}) = \mathcal{N}(\mathbf{0}, \mathbf{I})$ denote a base Gaussian, 
where $\mathbf{I}$ is the identity matrix with dimensionality matching that of $\boldsymbol{\theta}$. 
We then define a bijective, conditioning-dependent map 
$\boldsymbol{\theta} = f_{\varphi}(\mathbf{z}, \mathbf{h})$, 
parameterized by the neural-network weights $\varphi$.
By the change-of-variables formula, the conditional density is
\begin{equation}
p_{\varphi}(\boldsymbol{\theta}\,|\,\mathbf{h})
= p_{0}\!\left(f_{\varphi}^{-1}(\boldsymbol{\theta},\mathbf{h})\right)
  \left|
    \det 
    \frac{\partial f_{\varphi}^{-1}(\boldsymbol{\theta},\mathbf{h})}
         {\partial \boldsymbol{\theta}}
  \right|.
\end{equation}

We construct $f_{\varphi}$ as a composition of $K$ conditional affine coupling 
layers $f_k$ of the RealNVP/Glow family~\cite{Dinh2017,Kingma2018}:
\begin{equation}
f_{\varphi} = f_{K} \circ \cdots \circ f_{1}.
\end{equation}
Let $\mathbf{z}_0 = \mathbf{z}$ and 
$\mathbf{z}_{k} = f_{k}(\mathbf{z}_{k-1},\mathbf{h})$,
so that $\boldsymbol{\theta} = \mathbf{z}_K$.  
The log-density can then be written as
\begin{equation}
\log p_{\varphi}(\boldsymbol{\theta}\,|\,\mathbf{h})
= \log p_{0}(\mathbf{z}_0)
  - \sum_{k=1}^{K}
    \log
    \left|
      \det
      \frac{\partial f_{k}(\mathbf{z}_{k-1},\mathbf{h})}
           {\partial \mathbf{z}_{k-1}}
    \right|.
\end{equation}

Each coupling layer partitions the latent variable into two subsets
$\mathbf{z}=(\mathbf{z}_a,\mathbf{z}_b)$ and updates only $\mathbf{z}_b$:
\begin{align}
\mathbf{s}_{k},\,\mathbf{t}_{k} 
  &= g_{k}(\mathbf{z}_a, \mathbf{h}), \\
\mathbf{z}_b' 
  &= \mathbf{z}_b \odot \exp(\mathbf{s}_{k})
     + \mathbf{t}_{k}, \\
\mathbf{z}_a' 
  &= \mathbf{z}_a,
\end{align}
where $g_k$ is a small MLP conditioned on the context $\mathbf{h}$.  
The Jacobian of this transformation is triangular, so the 
log-determinant reduces to a simple sum:
\begin{equation}
\log\left|
\det \frac{\partial \mathbf{z}'}{\partial \mathbf{z}}
\right|
= \sum_i s_{k,i}.
\end{equation}
This yields exact and efficient evaluation of the likelihood, as well as 
exact inversion~\cite{Papamakarios2017}:
\begin{equation}
\mathbf{z}_b
= (\mathbf{z}_b' - \mathbf{t}_{k}) \odot \exp(-\mathbf{s}_{k}),
\qquad
\mathbf{z}_a = \mathbf{z}_a'.
\end{equation}

Training the DNF amounts to maximizing the conditional log-likelihood of the 
observed parameters given the context features.  
Given a dataset 
$\{(\boldsymbol{\theta}^{(n)},\mathbf{h}^{(n)})\}_{n=1}^{N}$, we minimize
\begin{equation}
\mathcal{L}_{\mathrm{DNF}}(\varphi)
= -\frac{1}{N}\sum_{n=1}^{N}
  \log p_{\varphi}(\boldsymbol{\theta}^{(n)}\,|\,\mathbf{h}^{(n)}).
\end{equation}
By construction, the DNF provides exact likelihoods, exact sampling, 
and an expressive, multi-modal posterior over astrophysical parameters.  
In our experiments, this DNF-based posterior estimator is the primary model 
used for quantitative results.

\subsubsection{Continuous Normalizing Flow:}
\label{subsubsec:cnf}

For completeness, we also consider a Continuous Normalizing Flow (CNF)
as an alternative invertible parameterization of the conditional posterior
\cite{Chen2018}. In this formulation, the flow is defined as the solution of an
ordinary differential equation,
\begin{equation}
\frac{d\boldsymbol{\theta}(t)}{dt}
= f_{\psi}\!\left(\boldsymbol{\theta}(t), \mathbf{h}, t\right),
\end{equation}
where $f_{\psi}$ is a neural network that specifies a time-dependent velocity
field in parameter space, conditioned on the physics-informed context vector
$\mathbf{h}$. The artificial time variable $t\in[0,1]$ parametrizes a continuous
deformation from a simple base distribution at $t=0$ to the target posterior at
$t=1$.

The evolution of the log-density along the flow is governed by the
instantaneous change-of-variables formula,
\begin{equation}
\frac{d}{dt}\log p(\boldsymbol{\theta}(t))
= -\mathrm{Tr}\!\left(
\frac{\partial f_{\psi}}{\partial \boldsymbol{\theta}}
\right),
\end{equation}
where the trace corresponds to the divergence of the vector field,
\(
\mathrm{Tr}\!\left(\partial f_{\psi}/\partial \boldsymbol{\theta}\right)
= \sum_{i=1}^{P} \partial f_{\psi,i}/\partial \theta_i
\),
with $P$ denoting the dimensionality of the inferred parameter vector
$\boldsymbol{\theta}$. This quantifies the local expansion or compression of probability mass induced by
the flow. In practice, this trace is estimated using a Hutchinson-type stochastic
trace estimator, which avoids explicit construction of Jacobian matrices
\cite{Hutchinson1990,Grathwohl2019}.

To evaluate the conditional log-likelihood during training, the ODE is
integrated in reverse time from $\boldsymbol{\theta}(1)=\boldsymbol{\theta}$ to
$\boldsymbol{\theta}(0)=\mathbf{z}$, where $\mathbf{z}$ follows a standard
Gaussian base density $p_0(\mathbf{z})=\mathcal{N}(\mathbf{0},\mathbf{I})$. This
yields
\begin{equation}
\log p(\boldsymbol{\theta}\mid\mathbf{h})
= \log p_{0}(\mathbf{z})
  - \int_{0}^{1}
    \mathrm{Tr}\!\left(
     \frac{\partial f_{\psi}}{\partial \boldsymbol{\theta}}
    \right)\,dt.
\end{equation}

While CNFs are typically more computationally expensive than discrete flows due
to the need for numerical ODE integration, they provide a smooth,
continuous-time parameterization of the transport map and impose fewer
architectural constraints. In this work, we employ CNFs primarily as a
complementary benchmark to assess the robustness of our DNF-based inference
results \cite{Papamakarios2019}. The conditioning network is evaluated in two phase-handling regimes:
\begin{enumerate}
    \item \textbf{No phase:} no phase information is provided to the model,
          and the phase-conditioned pathway is disabled, so that the
          conditioner relies only on positional and
          convolutional--Transformer features.
    \item \textbf{Predicted phase:} the model uses the phase sequence predicted
          by the pretrained phase-prediction network (PhaseProvider); the true
          phase is never exposed.
\end{enumerate}
This setup allows us to quantify the impact of predicted phase information on
the inferred posteriors relative to a phase-agnostic baseline.

Moreover, for the higher-dimensional inference in
Sec.~\ref{result:large:data}, we restrict the direct phase-conditioned correction to the orbital-evolution parameters $\boldsymbol{\theta}_{\phi}=(\log_{10}n_0,e_0,\log_{10}M)$. This choice provides a physics-motivated inductive bias, since these parameters most directly control the orbital-phase evolution, while avoiding unnecessary direct phase conditioning of the remaining parameters. The mask is applied to the phase-dependent scale and shift corrections in the DNF and to the phase-dependent component of the vector field in the CNF~\cite{Papamakarios:2017tec,gudovskiy2024contextflow++}. All parameters remain jointly inferred through the same flow, preserving cross-parameter correlations.

\begin{figure}[t]
    \centering
    \includegraphics[width=0.60\linewidth]{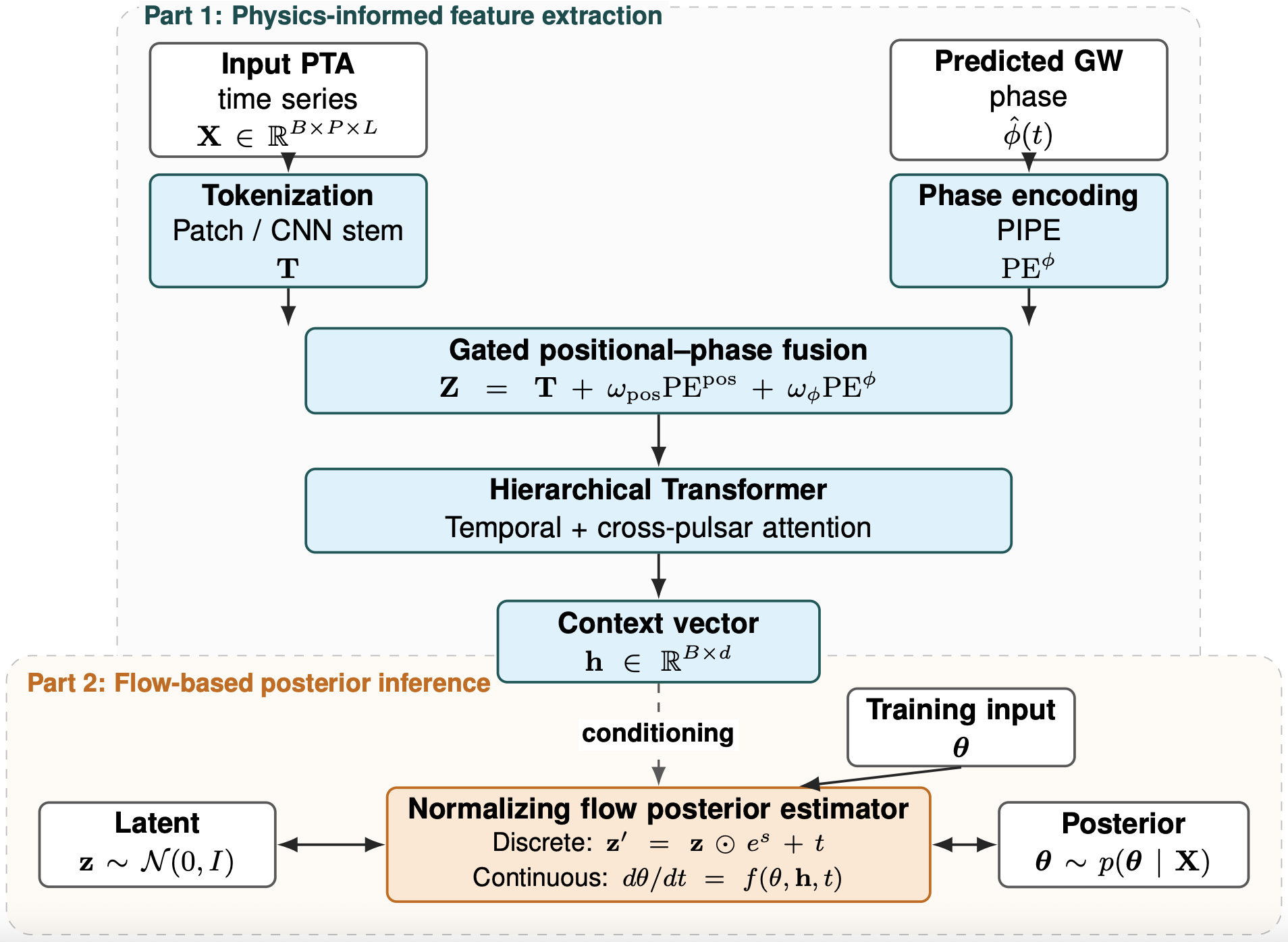}
    \caption{
Overall architecture of our physics-informed simulation-based inference pipeline
for EBBH systems in PTA data.
\textbf{Part 1: Feature extraction} (top). The PTA residual time series $X$ is
tokenized into features $T$, augmented with gated positional and orbital-phase
encodings, and processed by Transformer blocks to produce a
compact context vector $h$.
\textbf{Part 2: Inference} (bottom). A conditional normalizing flow $f_\varphi$
maps latent draws $z \sim \mathcal{N}(0,I)$ to posterior samples over the
physical EBBH parameters $\theta$, conditioned on the context $h$. We consider
both a discrete coupling-based flow and a continuous ODE-based flow, yielding an
efficient approximation to the posterior $p(\theta \mid X)$.
}
\label{fig:model-architecture}
    \label{fig:model-architecture}
\end{figure}

\section{Training}
\label{sec:training}
The simulated dataset is split into training and validation subsets, with $10\%$ reserved for validation (Table~\ref{tab:dataset-summary}). Input time series and target parameters are standardized using z-score normalization, with statistics computed from the training set and applied unchanged to validation data. Predicted parameters are transformed back to physical units using the corresponding inverse normalization.

Each sample consists of a noisy PTA timing-residual realisation and its associated EBBH parameters. Unless otherwise stated, we focus on the intrinsic subset
\begin{equation*}
 \boldsymbol{\theta}
=
\left(
\log_{10} n,\;
e_0,\;
\log_{10} M,\;
\log_{10}\mathcal{S}
\right),   
\end{equation*}
while performing inference over a higher-dimensional EBBH parameter space.

For phase-conditioned experiments, we first train a dedicated phase-prediction network in realisation mode. Each sample contains the full multi-pulsar time series $X \in \mathbb{R}^{P\times L}$ and a shared Earth-term GW phase trajectory $\phi(t)$, together with the corresponding realisation-level SNR. The phase model is trained on a deliberately large PTA simulation set comprising $4\times10^{5}$ realisations, each containing 10 pulsars, to promote robust phase recovery across the parameter space (as described in Sec.~\ref{sec:dataset}). To further improve robustness over signal strength, the realisation-level SNR is drawn log-uniformly,
\begin{equation*}
\log_{10}(\mathrm{SNR})
\sim
\mathcal{U}\!\left(\log_{10}10,\log_{10}100\right), 
\end{equation*}
and Gaussian noise is added according to the realisation-level white-noise SNR definition in Eq.~\eqref{eq:snr_realisation_wn}. The network jointly predicts the phase trajectory, represented by $(\cos\phi,\sin\phi)$, and $\log_{10}(\mathrm{SNR})$, thereby learning both temporal phase structure and global signal strength.

The phase predictor is implemented as a Transformer architecture with a convolutional downsampling stem of stride $S=4$, model dimension $D=128$, four encoder layers, four attention heads, feed-forward dimension $512$, and dropout $0.1$. It is trained for $100$ epochs with batch size $256$ using AdamW optimization with learning rate $2\times10^{-4}$, weight decay $10^{-4}$, gradient clipping at $1.0$, and a cosine-annealing learning-rate schedule. The training objective follows the multi-component phase and SNR loss defined in Sec.~\ref{subsec:phase_net}. After training, the phase predictor is frozen and used only to provide predicted phase trajectories for downstream inference.


The posterior conditioning network is a hierarchical Transformer
architecture. A per-pulsar temporal encoder first processes each pulsar
independently using patch embeddings followed by standard multi-head
self-attention Transformer blocks with latent dimension $128$, eight
attention heads, two temporal layers, feed-forward dimension $256$, and
patch size $20$. Standard positional encodings, and optionally predicted
phase encodings, are added at this stage. The resulting pulsar-level
embeddings are then passed to a two-layer cross-pulsar self-attention
encoder, which captures inter-pulsar correlations and produces a global
context vector $\mathbf{h}$ by mean pooling.

Both the Discrete Normalizing Flow (DNF) and Continuous Normalizing Flow (CNF) use this shared conditioning network and are trained by minimizing the negative conditional log-likelihood,
\begin{equation}
\mathcal{L}_{\mathrm{NLL}}
=
-\frac{1}{N_{\mathrm{batch}}}
\sum_{j=1}^{N_{\mathrm{batch}}}
\log q_\phi\!\left(\boldsymbol{\theta}_j \mid \mathbf{h}_j\right).
\end{equation}
The DNF uses a coupling-based architecture with affine coupling layers, whereas the CNF defines the transformation through a neural ODE with a shared vector field. Both models are optimized with AdamW using learning rate $2\times10^{-4}$, weight decay $10^{-4}$, gradient clipping at $1.0$, and batch size $128$. Training is performed for $120$ epochs, and the final checkpoint is selected by the minimum validation NLL.

We evaluate both \emph{no-phase} and \emph{predicted-phase} configurations. In the latter case, phase information is supplied by the pretrained phase provider and incorporated through phase-based encodings in the temporal encoder; the phase predictor remains fixed throughout posterior training. Table~\ref{tab:model_params} summarizes the parameter counts for the CNF and DNF models, separating the shared conditioner, frozen phase encoder, and flow components. The main architectural difference between CNF and DNF lies in the flow parameterization: CNFs use fewer flow parameters but require continuous-time density evaluation, while DNFs use a larger discrete coupling flow that is computationally more direct.

\begin{table*}[t]
\centering
\caption{
Total parameter counts for the CNF and DNF posterior models, with and without predicted phase conditioning, for the parameters listed in Table~\ref{tab:dataset-summary}. The phase encoder is pretrained and kept fixed; only the posterior conditioner and flow parameters are optimized during SBI training.
}
\label{tab:model_params}
\begin{tabular}{l c c c c}
\hline\hline
Model & Conditioner & Phase encoder & Flow & Total \\
\hline

DNF (no phase)
& $5.579\times10^{5}$
& --
& $2.028\times10^{5}$
& $7.607\times10^{5}$ \\

DNF (predicted phase)
& $1.116\times10^{6}$
& $9.754\times10^{5}$
& $4.711\times10^{5}$
& $2.562\times10^{6}$ \\

CNF (no phase)
& $5.579\times10^{5}$
& --
& $5.127\times10^{4}$
& $6.092\times10^{5}$ \\

CNF (predicted phase)
& $1.116\times10^{6}$
& $9.754\times10^{5}$
& $1.148\times10^{5}$
& $2.206\times10^{6}$ \\

\hline\hline
\end{tabular}
\end{table*}
\section{Evaluation Protocol}
\label{sec:evaluation}
We evaluate the proposed physics-informed simulation-based inference (SBI) framework using complementary criteria that assess posterior accuracy and reliability. Specifically, we consider posterior sharpness and calibration.

\subsection{Posterior Sharpness: Log Posterior Density at Truth}

To quantify how well the learned posterior concentrates around the true parameters, we evaluate the log posterior density at the ground truth,
\begin{equation}
\mathrm{LPD}(\boldsymbol{\theta}^\star) \equiv \log q_\phi(\boldsymbol{\theta}^\star \mid \mathbf{x}),
\end{equation}
where $q_\phi(\boldsymbol{\theta} \mid \mathbf{x})$ denotes the learned conditional posterior and $\boldsymbol{\theta}^\star$ are the true parameters. We report the mean LPD averaged over the validation set. Higher values indicate that the posterior assigns greater probability mass to the true parameters, reflecting both accuracy and sharpness. This metric is used to compare the impact of phase conditioning (predicted phase vs.\ no phase) across both CNF and DNF models.

\subsection{Calibration and Coverage}

We also assess posterior calibration using empirical coverage tests. For each parameter and credibility level $\alpha \in \{68\%, 95\%, 99.7\%\}$, we compute the fraction of validation samples for which the true value lies within the corresponding posterior credible interval,
\begin{equation}
\mathrm{Coverage}_\alpha = \frac{1}{N_{\mathrm{val}}} \sum_{n=1}^{N_{\mathrm{val}}}
\mathbf{1}\bigl( \boldsymbol{\theta}^{(n),\star} \in \mathcal{C}_\alpha^{(n)} \bigr),
\end{equation}
where $\mathcal{C}_\alpha^{(n)}$ denotes the $\alpha$-level credible region. A well-calibrated posterior satisfies $\mathrm{Coverage}_\alpha \approx \alpha$.

\section{Results}
\label{sec:results}
We infer the parameters of eccentric binary black hole (EBBH) systems from PTA noisy data using an amortized simulation-based inference (SBI) pipeline. The framework, illustrated in Figs.~\ref{fig:combined_phase_sbi_architecture} and ~\ref{fig:model-architecture}, consists of a physics-informed hierarchical Transformer encoder that maps a full PTA realisation $X \in \mathbb{R}^{P \times L}$ (with $P=10$ pulsars) into a latent context vector, followed by a normalizing flow that models the conditional posterior $q_\phi(\theta \mid X)$. All results are presented in realisation mode, where each sample corresponds to a full multi-pulsar observation sharing a common GW phase trajectory. Unless stated otherwise, we consider signals with $\mathrm{SNR} \in [20,30]$ and focus on the key intrinsic parameters $\{\log_{10} n,\; e_0,\; \log_{10} M,\; \log_{10} \mathcal{S}\}$, while performing higher-dimensional inference over the complete parameter set.

\subsection{Phase Prediction as Physics-Informed Conditioning}
\label{result:ph:pred}
We begin by validating the phase-prediction network, which provides the Earth-term GW phase used as an additional conditioning signal in the Transformer encoder. Two configurations are considered: (i) a baseline model using standard positional encoding (\emph{no phase}), and (ii) an augmented model incorporating a predicted phase encoding (\emph{pred phase}).

\begin{figure*}[t]
\centering
\includegraphics[width=\textwidth]{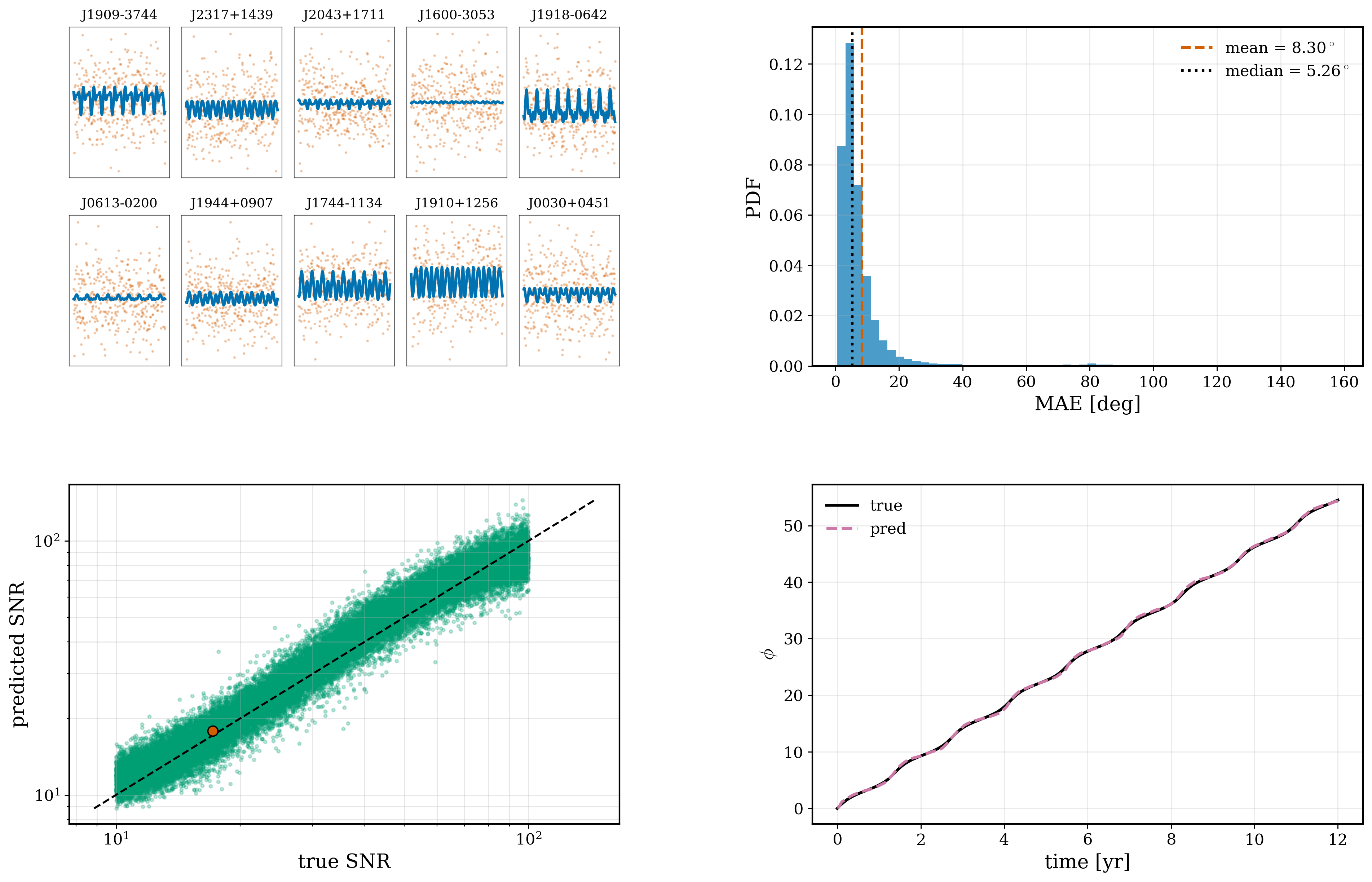}
\caption{Phase and SNR prediction performance of the phase-conditioned model evaluated on a representative validation sample. \textbf{Top-left:} Clean and noisy timing residuals for all pulsars in the array, showing the multi-pulsar realisation used as input.
\textbf{Top-right:} Distribution of phase alignment error across the full validation set, quantified by the mean absolute wrapped phase difference, $\langle |\Delta \phi| \rangle$, per sample, with mean and median indicated. \textbf{Bottom-left:} Predicted versus true realisation-level SNR (as defined in Eq.~\eqref{eq:snr_realisation_wn}) across the validation set, plotted on logarithmic ($\log_{10}$) axes. The dashed diagonal indicates perfect prediction, while the highlighted point marks the representative validation sample shown in the other panels (true $\mathrm{SNR}=17.2$, predicted $\mathrm{SNR}=17.9$). The model achieves strong agreement with an $R^2$ score of $\approx 0.93$ in linear space (and $\approx 0.96$ in $\log_{10}$ space).
\textbf{Bottom-right:} Reconstructed phase evolution for the representative validation sample highlighted by the red marker in the SNR plot, compared with the true phase in physical time units (years). The model accurately recovers the global phase trajectory despite noise.}
\label{fig:phase_snr_results}
\end{figure*}

As shown in Fig.~\ref{fig:phase_snr_results}, the phase model achieves strong performance across the validation set. The distribution of mean per-sample phase errors $\langle |\Delta\phi| \rangle$ is concentrated at low values, with a mean error of $8.30^\circ$ and a median error of $5.26^\circ$, demonstrating robust phase recovery at the population level. The predicted versus true realisation-level SNR (shown on logarithmic axes) exhibits a tight correlation along the identity line, with coefficient of determination $R^2 \approx 0.93$ in linear space and $R^2 \approx 0.96$ in $\log_{10}$ space. This indicates that the network reliably captures the underlying signal strength, which serves as a key conditioning variable for stable phase estimation.

To illustrate the phase prediction performance, we consider a randomly selected representative validation example, highlighted as a red dot in the SNR plot of Fig.~\ref{fig:phase_snr_results}. Even at a moderate SNR ($\sim17$), where the GW signal is significantly contaminated by noise, the model successfully recovers the global phase structure. For this example, the reconstructed phase $\hat{\phi}(t)$ closely tracks the true phase $\phi(t)$ with a mean absolute phase error (MAE) of $9.14^\circ$, and the corresponding realisation-level SNR is accurately estimated, with true and predicted values of $17.2$ and $17.9$, respectively. These results demonstrate that the predicted phase provides a robust and physically meaningful summary of the latent GW dynamics, and can therefore be used as a physics-informed conditioning signal in all downstream inference models.

\subsection{Posterior Inference with CNFs and DNFs}

\begin{figure*}[p]
\centering

\includegraphics[height=0.46\textheight,keepaspectratio]{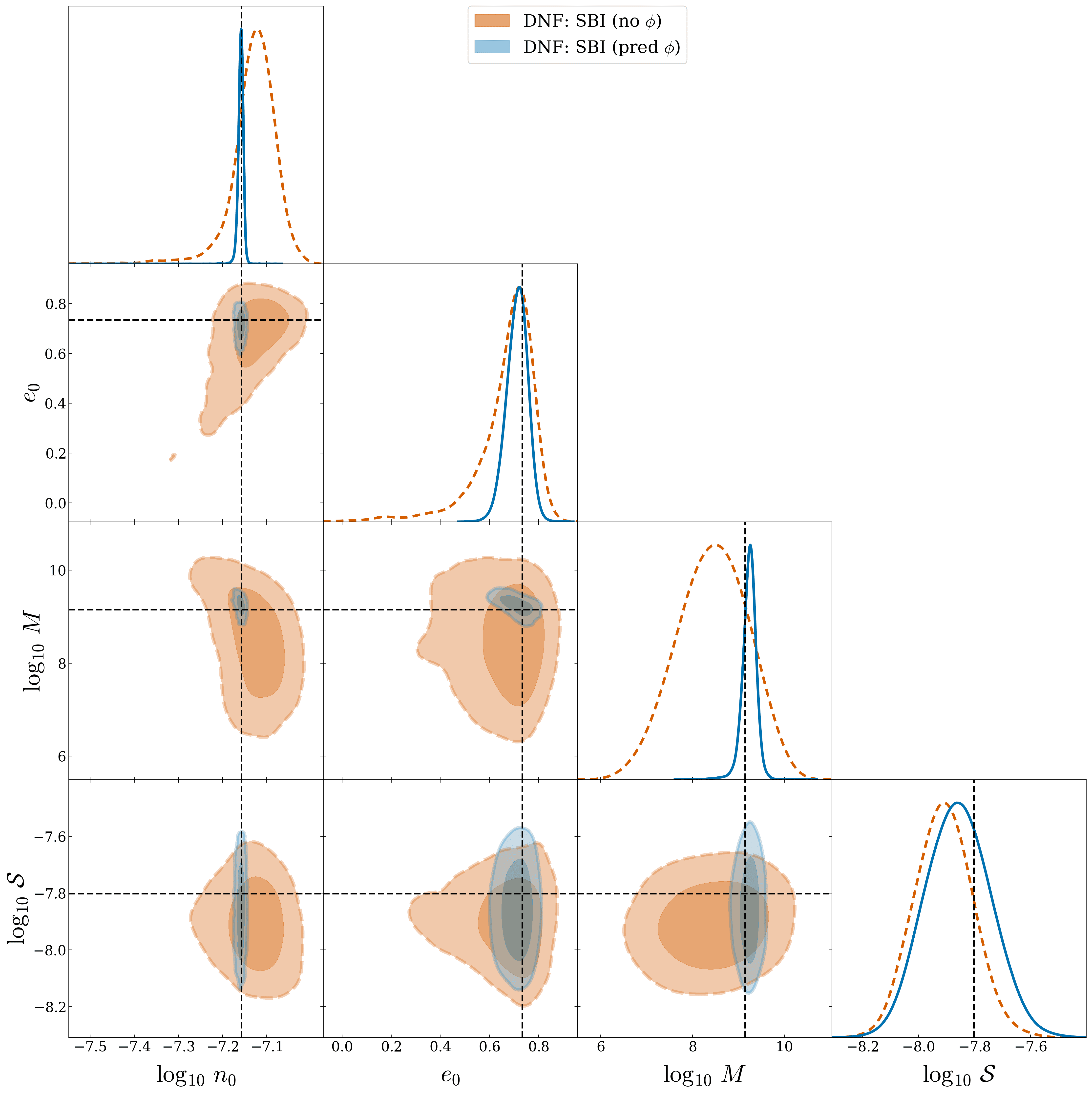}

\vspace{0.4em}

\includegraphics[height=0.46\textheight,keepaspectratio]{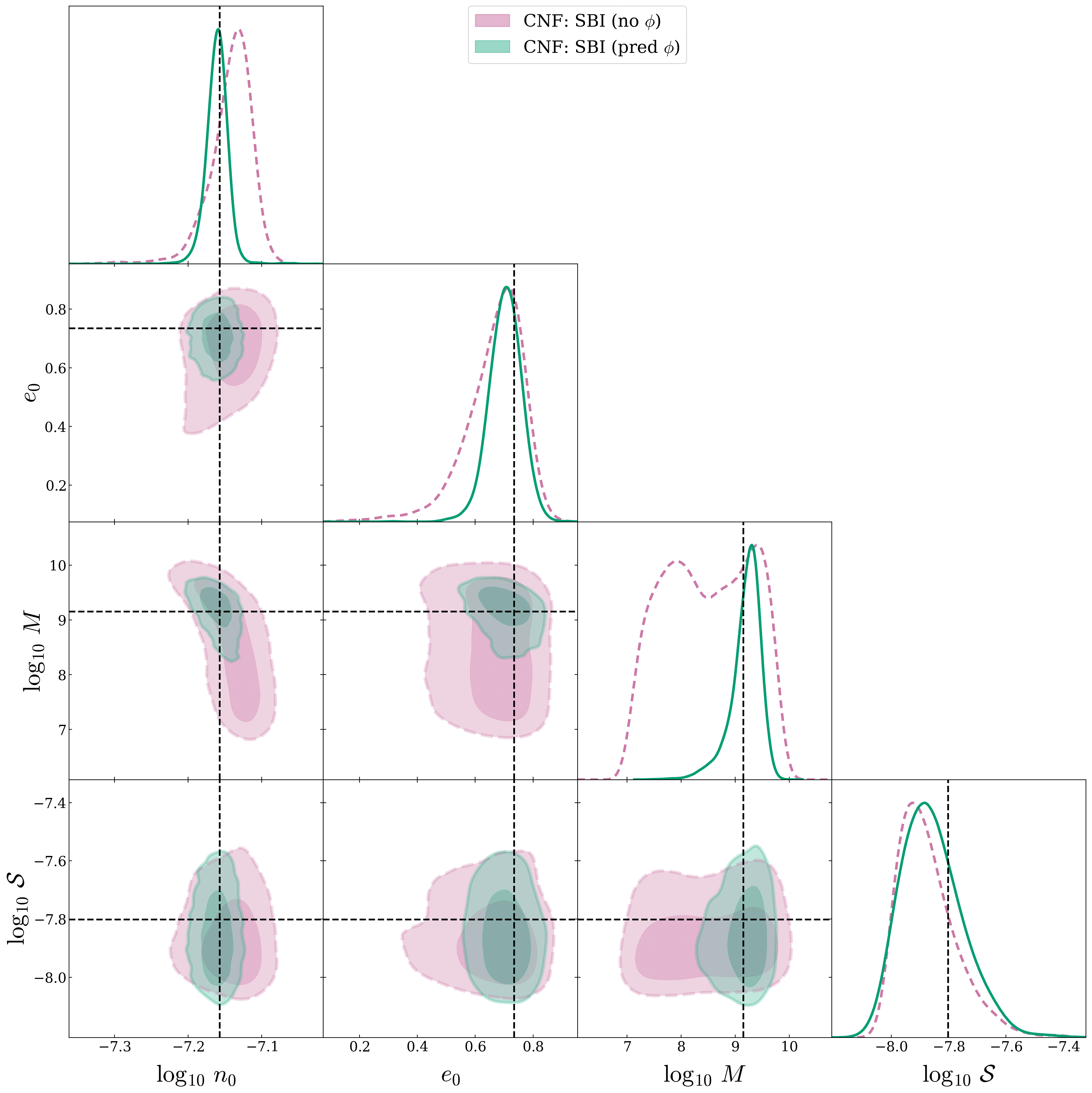}

\vspace{-0.5em}

\caption{\small
Posterior distributions for a representative validation sample with $\mathrm{SNR}\in[20,30]$, comparing no-phase and predicted-phase encodings for the DNF architecture (top) and CNF architecture (bottom). Posteriors are shown for $\{\log_{10} n, e_0, \log_{10}M, \log_{10}\mathcal{S}\}$, with $1\sigma$ and $2\sigma$ contours. Dashed black lines indicate the injected parameter values.
}
\label{fig:dnf_cnf_side_by_side}
\end{figure*}

We next examine the posterior distributions obtained using conditional normalizing flows. Fig.~\ref{fig:dnf_cnf_side_by_side} presents representative posterior estimates for both Continuous Normalizing Flows (CNFs) and Discrete Normalizing Flows (DNFs) under two phase configurations (\emph{no phase} and \emph{pred phase}) for a single validation realisation.

Across both flow families, the recovered posteriors are centered close to the true parameter values (indicated by dashed lines), demonstrating accurate amortized inference. Incorporating predicted phase information consistently sharpens the posterior distributions, leading to tighter credible regions and reduced degeneracies, particularly for parameters such as $\log_{10} n$, $e_0$, and $\log_{10} M$, which are strongly coupled to the orbital phase evolution. This indicates that the learned phase trajectories provide informative conditioning for downstream inference. However, it is worth noting that their contribution is modulated through the learned phase-encoding weights defined in Eq.~\eqref{eq:encoding-weight-gradient}, allowing the model to reduce the influence of the phase channel when the predicted phase is less accurate.

While both CNFs and DNFs achieve comparable posterior accuracy, CNFs incur a higher computational cost (approximately $\sim\!3\times$ in our scenario) due to the need to solve continuous-time dynamics during density evaluation. In contrast, DNFs provide a more computationally efficient alternative, making them preferable in scenarios where fast inference is required.

\subsection{Log Posterior Density at Truth}

Table~\ref{tab:lpd_comparison_realisation} reports the mean exact log posterior density (LPD) evaluated at the ground-truth parameters, averaged over validation realisations. Higher values indicate that the learned posterior assigns greater density to the ground-truth parameters.

We observe a substantial improvement when incorporating predicted phase information. Relative to the no-phase baselines, the LPD increases from $-0.715$ to $0.650$ for the CNF and from $-1.197$ to $0.805$ for the DNF, indicating that phase conditioning substantially improves posterior fidelity by assigning higher probability density to the true parameters. While the CNF outperforms the DNF in the absence of phase conditioning, the performance gap narrows considerably once predicted phase information is incorporated, suggesting that informative physical conditioning plays a more important role than the specific choice of flow architecture.
\begin{table}[t]
\centering
\caption{Mean exact log posterior density (LPD) evaluated at the ground-truth parameters
in the standardized parameter space, averaged over validation realisations from
the dataset described in Table~\ref{tab:dataset-summary}. All values are computed from the exact model log density, $\log q_\phi(\theta^\star \mid x)$, in the standardized parameter space. Higher LPD values indicate that the learned posterior assigns greater probability density to the true parameter values.}
\label{tab:lpd_comparison_realisation}
\begin{tabular}{l c}
\hline\hline
Model & $\langle \log q_\phi(\theta^\star \mid x) \rangle$ \\ 
\hline
CNF (no phase)            & $-0.715$ \\
CNF (predicted phase)    & $0.650$ \\
DNF (no phase)            & $-1.197$ \\
DNF (predicted phase)    & $0.805$ \\
\hline\hline
\end{tabular}

\end{table}
\subsection{Large-Data Regime Behavior}
\label{result:large:data}
\begin{figure*}[!tp]
    \centering
    \includegraphics[width=\textwidth]{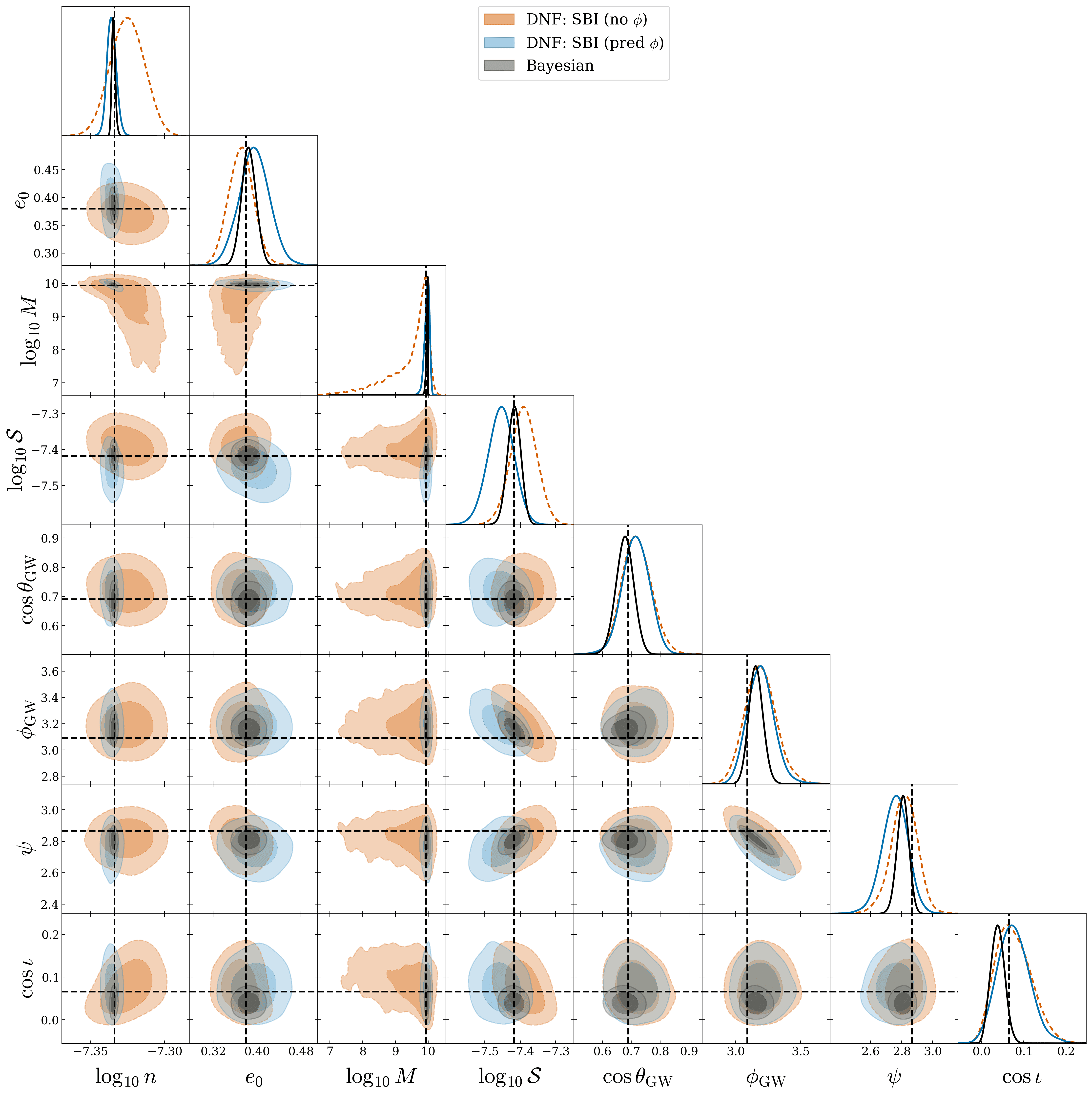}
    \caption{
Higher-dimensional posterior distributions over the EBBH parameter space for a representative validation sample, comparing DNF-based SBI models without phase encoding (no $\phi$), with predicted phase encoding (pred $\phi$), and Bayesian inference. The true parameter values are indicated by black dashed lines. The model is trained on a large PTA realisation dataset ($\sim 4\times10^{5}$ samples, each comprising 10 pulsars). In this data-rich regime, all methods yield broadly consistent posteriors, with predicted phase encoding providing modest sharpening of constraints for some parameters (notably $\log_{10} n$, $e_0$, and $\log_{10} M$), while overall differences remain small.
}
    \label{fig:corner_full_posterior_large_data}
\end{figure*}

As described in Sec.~\ref{sec:dataset}, this analysis uses an expanded training dataset comprising $4\times10^5$ realisations, in contrast to the default dataset used in the preceding inference experiments. This configuration allows us to assess the behavior of the amortized SBI posterior in the data-rich regime and to compare it directly with posteriors obtained from conventional full Bayesian inference over the same EBBH parameter space. Fig.~\ref{fig:corner_full_posterior_large_data} shows posterior distributions for a representative validation sample drawn from a held-out set comprising $10\%$ of the expanded dataset, in the large-data regime. In this scenario, the \emph{no phase} and \emph{pred phase} configurations yield highly consistent posteriors, with only modest differences in marginal and joint constraints.

This behavior suggests that, given sufficient training data, the model can effectively learn phase-dependent structure directly from the data, thereby reducing the incremental benefit of explicit phase conditioning. The most noticeable differences are observed in parameters such as $\log_{10} n$ and $\log_{10} M$, where predicted phase encoding leads to slight sharpening of the posterior.
Overall, this highlights a key trade-off: physics-informed conditioning provides the greatest advantage in data-limited regimes.
\subsection{Coverage and Calibration}
\begin{figure*}[t]
    \centering
    \includegraphics[width=\textwidth]{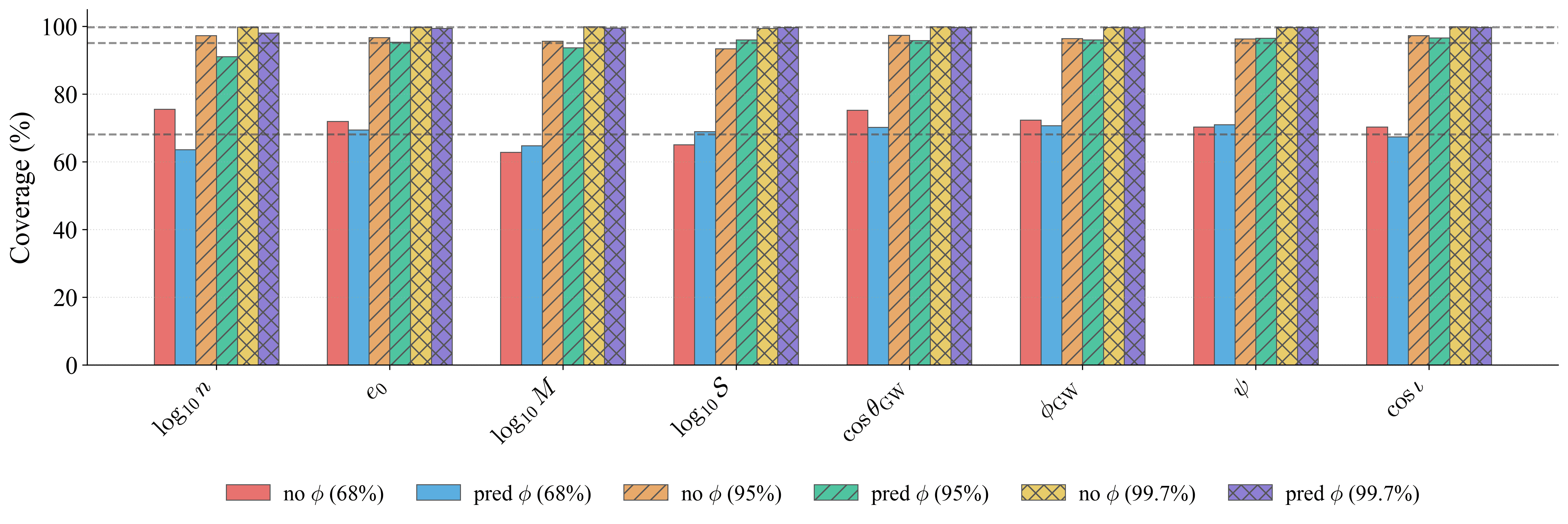}
    \caption{
Marginal coverage comparison for simulation-based inference (DNFs) using
no phase encoding (no $\phi$) and predicted phase encoding (pred $\phi$), evaluated at
nominal credibility levels of $68\%$, $95\%$, and $99.7\%$
($1\sigma$, $2\sigma$, and $3\sigma$), over the validation set
(10\%) of the expanded PTA dataset comprising $\sim4\times10^5$
realisations (10 pulsars per realisation) with
$\mathrm{SNR}\in[20,30]$. Dashed horizontal lines indicate the ideal expected coverage at each level. While both methods achieve near-nominal coverage at $99.7\%$, larger deviations 
are observed at $68\%$, where predicted phase encoding improves calibration 
for several parameters, although performance remains parameter-dependent.
}
    \label{fig:coverage_comparison}
\end{figure*}

We further assess posterior calibration over the same higher-dimensional held-out dataset considered in Sec.~\ref{result:large:data}. Marginal coverage
probabilities are shown in Fig.~\ref{fig:coverage_comparison}. Predicted-phase conditioning improves the overall calibration, reducing the coverage MAE from $2.09\%$ to $1.35\%$ relative to the no-phase
baseline.

The largest improvement occurs at the $68\%$ credibility level, where explicit phase information reduces deviations from nominal coverage for several
parameters. At the $99.7\%$ level, both variants remain close to nominal coverage across the inferred parameter space. However, these improvements remain parameter-dependent, reflecting varying sensitivity to phase information.

\section{Conclusion and Future Directions}
\label{sec:conclusion}
We have introduced a physics-informed simulation-based inference framework for parameter estimation of deterministic EBBH signals in pulsar timing-array data. Although developed here for EBBH systems, the framework is broadly applicable to other deterministic PTA sources with structured temporal evolution. The proposed pipeline combines a hierarchical physics-informed Transformer encoder with conditional normalizing flows, enabling amortized posterior inference from full multi-pulsar PTA realisations. Unlike purely data-driven sequence models, the Transformer representation is augmented with physics-informed positional encodings derived from the orbital phase evolution of eccentric binaries. This provides the network with an explicit inductive bias toward the dynamical structure of the underlying GW signal.

The framework operates at the realisation level, where all pulsars associated with a common source, namely the EBBH system, are processed jointly as a full PTA observation. A temporal encoder first extracts within-pulsar features from the residual time series, while a cross-pulsar attention module aggregates array-level information into a compact context vector. This context conditions a normalizing-flow posterior estimator, allowing efficient sampling from $q_\phi(\theta \mid X)$ and providing uncertainty estimates, parameter correlations, and degeneracies. We considered both discrete coupling-based normalizing flows and continuous normalizing flows. While both flow architectures yield comparable posterior quality, the discrete normalizing flow is computationally more efficient in our analysis and is therefore better suited for fast amortized inference.

A key component of the proposed method is the phase-prediction network, which estimates the shared Earth-term orbital phase directly from noisy multi-pulsar residuals, since this quantity is not directly observable in PTA data. The model recovers the global phase trajectory accurately across the validation set and also provides reliable realisation-level SNR estimates. When used as an additional physics-informed conditioning signal, the predicted phase improves posterior sharpness and calibration in several cases, particularly for parameters closely tied to orbital evolution, such as $\log_{10} n$, $e_0$, and $\log_{10} M$. The improvement is most evident in the log-posterior density at the true parameters, where phase-conditioned models assign substantially higher density to the injected values than their phase-agnostic counterparts.

At the same time, the large-data experiments show an expected behaviour of physics-informed machine learning. When sufficient training data are available, both the data-driven baseline and the physics-informed model can learn the dominant phase-dependent structure, leading to broadly consistent posteriors. In this data-rich regime, explicit phase conditioning provides only modest additional sharpening rather than a qualitatively different inference outcome. Nevertheless, the phase-informed encoding remains valuable in data-limited regimes, lower-SNR scenarios, and extrapolative cases. The learned gating mechanism also acts as an adaptive safeguard, allowing the network to emphasize the phase channel when the predicted phase is reliable and down-weight it when the phase prediction is poor or noisy.

Notably, the present study focuses primarily on deterministic EBBH signals under a controlled white-noise approximation. This simplified formulation allows us to isolate the effect of physics-informed sequence modeling and posterior density estimation before moving to more realistic PTA analyses. A natural next step is to extend the framework to a full PTA noise model, including intrinsic pulsar achromatic red noise~\citep{lentati2016spin}, potential chromatic dispersion-measure variations~\citep{you2007dispersion, van_haasteren_measuring_2009}, and spatially correlated Hellings--Downs~\citep{hellings1983upper} components associated with a stochastic gravitational-wave background~\citep{agazie2023ipta3p+}.

In this higher-dimensional regime, physics-informed encodings could be extended beyond the EBBH orbital phase to represent both deterministic signal structure and stochastic noise structure. By incorporating pulsar-specific noise properties, chromatic propagation effects, and array-level spatial correlations such as Hellings--Downs correlations, such encodings could help the model infer a broader parameter space spanning deterministic source parameters, pulsar-noise parameters, and stochastic-background parameters. At the same time, when the primary scientific target is the deterministic EBBH source, the posterior estimator can marginalize over nuisance noise and background parameters, yielding calibrated constraints on the EBBH parameters alone. This extension would move the framework closer to realistic end-to-end amortized inference for PTA datasets.

Finally, while this work uses normalizing flows as the posterior estimator, the modular design allows the density-estimation component to be replaced by alternative generative models. Diffusion-based posterior~\citep{batzolis2021conditional,song2020score} samplers are a promising future direction, particularly for high-dimensional, multi-modal, or strongly non-Gaussian posteriors. In such a variant, the same physics-informed Transformer context vector could condition a diffusion model that iteratively denoises samples in parameter space, providing a flexible alternative to flow-based transport maps. Comparing flow-based and diffusion-based SBI in PTA inference would help clarify the trade-off between sampling speed, posterior expressivity, and calibration.

Overall, this work demonstrates that physics-aware Transformer architectures, combined with amortized neural posterior estimation, provide a scalable route toward rapid Bayesian-like inference for deterministic PTA signals such as eccentric binary black holes. The results indicate that physically structured encodings can improve posterior quality while retaining the flexibility of data-driven representation learning. With further extensions to realistic noise models, correlated backgrounds, and higher-dimensional source populations, the proposed framework may offer a practical complement to conventional Bayesian pipelines for next-generation GW analyses.


\ack{This work was supported by the Singapore National Research Foundation's AI4S Catalytic Grant AI4SCT\_2025-0020, and by the National University of Singapore (Grant No. A8000554-02-00). Computational resources, including GPU allocations, were provided by the National University of Singapore High Performance Computing (NUS HPC) Centre under the CFP03-CF-051 Project Resource Allocation Award. We thank Liu Miaoxin, Lankeswar Dey, and A. Gopakumar for useful discussions and valuable suggestions.}





\section*{Data availability statement}

The data that support the findings of this study are openly available at the
following DOI:
\url{https://doi.org/10.5281/zenodo.22972338}.

\section*{Code availability}

All source code is openly available at
\url{https://github.com/subhajitphy/PIPE-GWs-PTA} under the MIT License,
including the data-generation scripts, phase-prediction model, hierarchical
Transformer encoders, and DNF/CNF posterior-inference and training scripts.

\section*{Conflict of interest}

The authors declare no competing interests.

\bibliographystyle{iopart-num}
\bibliography{ref}

\end{document}